\documentclass[twocolumn]{svjour3}          
\smartqed  
\usepackage{bbm}

\usepackage{amsmath}
\usepackage{graphicx}
\usepackage{amssymb} 
\usepackage{bm}
\usepackage{color}
\usepackage{booktabs}
\usepackage{multirow}
\usepackage{rotating}
\usepackage{subfigure}
\usepackage{soul}
\usepackage{todonotes}
\usepackage{amsfonts}
\usepackage{natbib}
\usepackage{hyperref}
\usepackage[modulo,switch]{lineno}
\hypersetup{colorlinks=true,citecolor=blue,linkcolor=blue}

\newcommand{\scal}[1]{\mathit{#1}}
\newcommand{\vect}[1]{\mathbf{#1}}
\newcommand{\matr}[1]{\mathbf{#1}}
\begin{document}

\title{{Transductive Zero-Shot Action Recognition by Word-Vector Embedding}}


\author{Xun Xu         \and
        Timothy Hospedales \and 
        Shaogang Gong
}


\institute{Xun Xu, Timothy Hospedales and Shaogang Gong \at
              Queen Mary, University of London \\
              \email{xun.xu@qmul.ac.uk}           
           \and
           Timothy Hospedales \at
              \email{t.hospedales@qmul.ac.uk}
           \and
           Shaogang Gong \at
              \email{s.gong@qmul.ac.uk}
}

\date{Received: date / Accepted: date}

\maketitle

\begin{abstract}
The number of categories for action recognition is growing rapidly and
it has become increasingly hard to label sufficient training data
for learning conventional models for all  
categories. Instead of collecting ever more  data and
labelling them exhaustively for all  categories, an attractive
alternative approach is ``zero-shot learning'' 
(ZSL). To that end, in this study we construct a mapping between visual
features and a semantic descriptor of each action category, allowing new
categories to be recognised in the absence of any visual training
data. Existing ZSL studies focus primarily on still images, and
attribute-based semantic representations. In this work, we explore word-vectors as the shared semantic space to embed videos and
category labels for ZSL action recognition. This is a more challenging problem than existing ZSL
of still images and/or attributes, because the mapping between
video space-time features of actions and the semantic space is more
complex and harder to learn for the purpose of generalising over any
cross-category domain shift. {To solve this generalisation problem in
ZSL action recognition, we investigate a series of synergistic
strategies to improve upon the standard ZSL pipeline. Most of these strategies are transductive in nature which means access to testing data in the training phase.} First, we 
enhance significantly the semantic space mapping by proposing manifold-regularized regression and data
augmentation strategies. Second,
we evaluate two existing post processing strategies (transductive
self-training and hubness correction), and show that they are
complementary. We evaluate extensively our model on a wide range of human
action datasets including HMDB51, UCF101, OlympicSports and event datasets including CCV
and TRECVID MED 13. The results demonstrate that our approach achieves
the state-of-the-art zero-shot action recognition performance with a
simple and efficient pipeline, and without supervised
annotation of attributes. Finally, we present in-depth analysis into why and when zero-shot works, including demonstrating the ability to predict cross-category transferability in advance.

\end{abstract}

\section{Introduction}
\label{sec:intro}

Action recognition is of established importance in the computer vision community due to its potential applications in video retrieval, surveillance and human machine interaction \citep{Aggarwal_ActivityReview_2011}. However the need for increasing coverage and finer classification of human actions means the number and complexity of action categories of interest for recognition is growing rapidly. For example, action recognition dataset size and number of categories has experienced constant growth since the classic KTH Dataset \citep{Schuldt_KTH} (6 classes, 2004): Weizmann Dataset \citep{Blank_Weizmann_ICCV05} (9 classes, 2005), Hollywood2 Dataset \citep{Marszalek_Hollywood2_CVPR09} (12 classes, 2009), Olympic Sports Dataset \citep{NieblesCL_eccv10} (16 classes, 2010), HMDB51 \citep{Kuehne2011} (51 classes, 2011) and UCF101 \citep{Soomro2012} (101 classes, 2012). 
The growing number and complexity of actions result in: (1) Enormous
human effort is required to collect and label large quantities of video
data for  learning. Moreover, compared
to image annotation, obtaining each annotated action clip is more
costly as it typically requires some level of spatio-temporal
segmentation from the annotator. (2) The growing number of categories
eventually begins to pose ontological difficulty, about how to
structure and define distinct action categories as they grow more
fine-grained and inter-related \citep{DBLP:journals/corr/JiangWWXC15}. 
In this work, we explore methods which do
not explicitly create models for new action categories from manually
annotated training data, but rather dynamically construct recognition
models by combining past experience in language together with
knowledge transferred from already labelled existing action categories.   

The ``zero-shot learning'' (ZSL) paradigm
\citep{Lampert2009,fu2012attribute,Socher2013} addresses this goal by
sharing information across categories; and crucially by allowing
recognisers for novel/ unseen/ testing categories\footnote{We use {\em known}, {\em seen} and {\em training} interchangeably to refer to the categories with labeled visual training examples and {\em novel}, {\em unseen} and {\em testing} interchangeably to refer to the categories to be recognized without any labeled training samples.} to be constructed based on a semantic
{\em description} of the category, without any labelled {\em visual} training
samples. ZSL methods follow the template of learning a general
mapping between a visual feature and semantic descriptor space from known/ seen/ training
data . In the context of zero-shot action recognition, `semantic
descriptor' refers to an action class description
that can be specified by a human user, either manually, or with
reference to existing knowledge bases, e.g. wikipedia. The ZSL
paradigm is most commonly realised by using class-attribute descriptors
\citep{Lampert2014,Liu2011,Fu2015} to bridge the semantic gap between
low-level features (e.g. MBH or SIFT) and categories. Attributes are
mid-level concepts that transcend class boundaries \citep{Lampert2009},
allowing each category or instance to be represented as a binary
\citep{Lampert2009,Liu2011} or continuous \citep{Fu2014a}
vector. Visual attribute classifiers are learned for a set of known
categories, and then a human can create recognisers for novel
categories by specifying their attributes. 
With a few exceptions \citep{Liu2011,Fu2015,Xu_SES_ICIP15}, this
paradigm has been applied to images rather than video action
recognition.  

An emerging alternative to attribute-based ZSL is {\em unsupervised} semantic embeddings \citep{Socher2013,Frome2013,Fu2014a,Habibian2014,norouzi2014_ICLR,fu2015zero,Xu_SES_ICIP15,akata2015outputEmbedding}.Unsupervised 
semantic embedding spaces refer to intermediate representations which can
be automatically constructed from existing unstructured
knowledge-bases (such as wikipedia text), rather than manually
specified attributes. The most common approaches
\citep{Socher2013,Fu2014a,fu2015zero,Xu_SES_ICIP15,akata2015outputEmbedding}
are to exploit a distributed vector representation of words
produced by a  neural network \citep{Mikolov2013} trained on a large
text corpus in an unsupervised manner. Regressors (cf classifiers in the attribute space),
are trained on the known dataset to map low-level visual features into this semantic embedding space. Zero-shot recognition is subsequently performed by mapping novel
category visual instances to the embedding space via the regression, and matching
these to the vector representation of novel class names (e.g. by nearest
neighbour).  
Several properties make the embedding space approaches preferable to the attribute-based
ones: (1) A manually pre-defined attribute ontology is not needed as embedding space is learned in an unsupervised manner. (2) Novel categories can be defined trivially by
\emph{naming} them, without the requirement to exhaustively define
each class in terms of a list of attributes -- which grows
non-scale-ably as the breadth of classes to recognise grows
\citep{Fu2014a,akata2015outputEmbedding}. (3) Semantic embedding allows easier
exploitation of information sharing across datasets
\citep{Xu_SES_ICIP15,Habibian2014} because category names from multiple
datasets can be easily projected into a common embedding space, while attribute
spaces are usually dataset specific, with datasets having incompatible
attribute schemas 
(e.g. UCF101 \citep{THUMOS13}  and Olympic Sports \citep{Liu2011} have
disjoint attribute sets). 

\vspace{0.1cm}\noindent\textbf{The domain shift problem for ZSL of actions}\quad 
Although embedding-based ZSL is an attractive paradigm, it has rarely
previously been demonstrated  in zero-shot action
recognition. This is in part because of the pervasive challenge of
learning mappings, that generalize across
the train-test semantic gap \citep{Fu2015,paredes2015ez_zsl}. 
In ZSL, the train-test gap is more significant than conventional
supervised learning because the training and testing classes are
\emph{disjoint}, i.e. completely different without any overlap. As a result of
serious \emph{domain-shift} 
\citep{Pan_TransferSurvey_KDE10}, mapping from low-level visual feature to semantic embedding trained on
a known class data will generalise poorly to novel class data. This is because the
data distributions for the underlying categories are different. This
violates the assumptions of supervised learning methods and results
in poor performance. The domain shift problem -- analysed empirically in
\citet{Fu2015,dinu2015improving}, and theoretically in
\citet{paredes2015ez_zsl} -- is worse for action than still image
recognition because of the greater complexity of categories in
visual space-time features and
the mapping of space-time features to semantic embedding space. 

\begin{figure*}[!hbt]
\begin{center}
\includegraphics[width=0.95\linewidth]{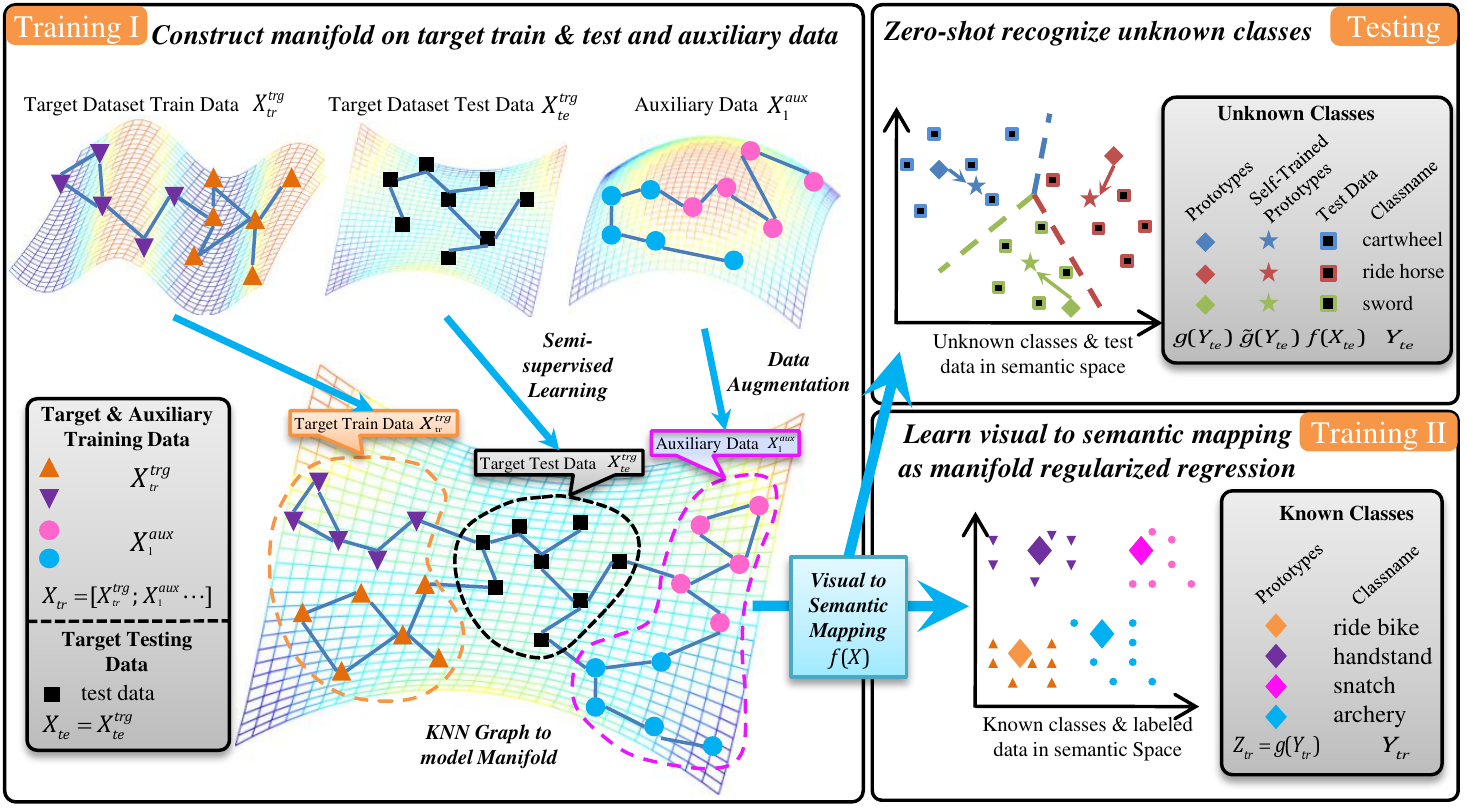}
\end{center}
\caption{{We have labelled data in target dataset $\matr{X}^{trg}_{tr}$ and auxiliary dataset $\matr{X}^{aux}_1$ and testing data in target dataset $\matr{X}^{trg}_{te}$. The objective is to use all this data to classify testing data into a set of pre-defined categories (aka unknown classes). Specifically, in the training phase I, target labelled data $\matr{X}^{trg}_{tr}$ is first augmented by data from auxiliary dataset $\matr{X}^{aux}_1$ to form a combined labelled dataset $\matr{X}_{tr}$. We construct a K Nearest Neighbour (KNN) graph on all labelled and testing data in visual feature space to model the underlying manifold structure. In the training phase II, prototypes for known classes are generated by semantic embedding $\matr{Z}_{tr}=g(\vect{y}_{tr})$. Then we learn a visual-to-semantic mapping $f:\matr{X}_{tr}\to \matr{Z}_{tr}$ as manifold regularized regression. In the testing phase, prototypes for unknown classes are first generated by semantic embedding $g(\vect{y}_{te})$. Then target testing data $\matr{X}_{te}$ are projected into semantic space via $f(\matr{X})$. Finally simple nearest neighbour (NN) classifier is used to categorize testing data as the label of closest prototype. On top of NN classifier, self-training and hubness corrections are adopted at testing phase to  improve results by mitigating the domain shift problem. With this framework we achieve the state-of-the-art performance on zero-shot action recognition tasks.}}
\label{fig:Pipeline}
\end{figure*}

\vspace{0.1cm}\noindent\textbf{Our Solutions}\quad 
In this work, we explore four potential solutions to ameliorate the
domain shift challenge in ZSL for action recognition as shown in Fig.~\ref{fig:Pipeline}, and achieve
better zero-shot action recognition: (1) The first strategy we consider aims to improve the generalisation of the
embedding space mapping. We explore \textbf{manifold regularization} (aka semi-supervised learning) to
learn a regressor which exploits a regularizer based on the testing/unlabelled
data to learn a smoother regressor that better generalises to novel
testing classes. Manifold regularization \citep{belkin2006manifold} is
established in semi-supervised learning to improve generalisation of
predictions on testing data, but this is more important in ZSL since the
gap between training and testing data is even bigger due to disjoint
categories. {To our best knowledge, this is the first transductive use of testing/unlabelled data for zero-shot learning at training time.} (2) The second strategy we consider is
\textbf{data augmentation}\footnote{{`Data augmentation' in this context means including data from additional datasets; in contrast to its usage in deep learning which refers to synthesising training examples by e.g. rotating and scaling.}} (aka {cross-dataset transfer learning})
\citep{Pan_TransferSurvey_KDE10,shao2015transfer}. The idea is that by
simultaneously learning the regressors for multiple action datasets, a
more representative sample of input action data is seen, and thus a
more generalizable mapping from the visual feature to the semantic embedding
space is learned. This is straightforward to achieve with semantic embedding-based ZSL because the datasets
and their category name word-vectors can be directly aggregated. In
contrast, it is non-trivial with attribute-based ZSL due to the need
to develop a universal attribute ontology for all datasets.
Besides these two new considerations to expand the embedding
projection, we also evaluate two existing post-processing heuristics
to reduce the effect of domain-shift in ZSL. These include
(3) \textbf{self-training}, which adapts test-class descriptors based on
unlabeled testing data to bridge the domain shift \citep{Fu2014b} and
(4) \textbf{Hubness correction} which re-ranks the test-data's match to
novel class descriptions in order to avoid the bias toward `hub'
categories induced by domain shift
\citep{dinu2015improving}.

By exploring manifold regularization, data augmentation, 
self-training, and hubness correction, our word-vector embedding
approach outperforms consistently conventional  zero-shot approaches
on all contemporary action datasets (HMDB51, UCF101, Olympic Sports,
CCV and USAA). \textcolor{black}{On a more relaxed
  multi-shot setting}, our representation is comparable with using
 low-level features directly. {Interestingly, with unsupervised semantic embedding (word-vector) and transductive access to testing data we are able to achieve very competitive performance even compared to supervised embedding methods 
 \citep{Fu2014a,akata2015outputEmbedding}  which  require attribute annotation. Moreover, because our method has a closed-form solution to the visual to semantic space mapping, it is very simple to implement, requiring only a few lines of Matlab.}

\vspace{0.1cm}\noindent{\textbf{Transductive Setting}}\quad {Of the four strategies, manifold regularization,  
self-training, and hubness correction assume access to the full set of unlabelled testing data, which is called the transductive setting \citep{belkin2006manifold,Fu2015}. This assumption would be true in many real-world problems. Video repositories, e.g. YouTube, can process large batches of unlabelled videos uploaded by users. Transductive zero-shot methods can be used to tag batches automatically without manual annotation, or add a new tag to the ontology of an existing annotated set.}

\vspace{0.1cm}\noindent{\textbf{New Insights}}\quad
In order to better understand ZSL, this study
performs a detailed analysis of the relationship between training and
testing classes for zero-shot learning, revealing the causal connection
between known and novel category recognition performance. 

\vspace{0.1cm}\noindent\textbf{Contributions}\quad Our key
contributions are threefold: (1) We explore jointly four mechanisms for
expanding ZSL by addressing its domain-shift challenge, {including three transductive learning strategies - manifold regularization, self-training and hubness correction}. Our model is both {\em closed-form} in solving the visual to semantic mapping and {\em
  unsupervised} in constructing the semantic embeddings. (2) We show extensive experiments to demonstrate a
very simple implementation of this closed-form model that both runs
very quickly and is capable of achieving the state-of-the-art ZSL
performance on contemporary action/event datasets. (3) We provide
new insight, for the first time, into the underlying factors affecting
the efficacy of ZSL.


\section{Related Work}

\subsection{Action Recognition}
Video action recognition is now a vast and established area in
computer vision and pattern recognition due to the wide application in
video surveillance, interaction between human and electronic
devices. Extensive surveys of this area are conducted by
\citet{Aggarwal_ActivityReview_2011,poppe2010actionrec_survey}. Recent
progress in this area is attributed to densely tracking points and
computing hand-crafted features which are fed into classical supervised
classifiers (e.g. SVM) for recognition \citep{WangOVS_IJCV15}. 

\vspace{0.1cm}\noindent\textbf{Human Action Datasets}\quad Video
datasets for action recognition analysis have experienced constant
developing. Early datasets focus on simple and isolated human actions
performed by a single person, e.g. KTH \citep{Schuldt_KTH} (2004)
 and Weizmann \citep{Blank_Weizmann_ICCV05} (2005)  datasets. Due to
the growth of internet video sharing, e.g. YouTube and
Vimeo, action datasets collected from online repositories are
emerging, e.g. OlympicSports \citep{NieblesCL_eccv10} in 2010,
HMDB51 \citep{Kuehne2011} in 2011 and UCF101 \citep{Soomro2012} in
2012.

\vspace{0.1cm}\noindent\textbf{Event Datasets} To recognize more complex events with interactions
between people and objects, event datasets including Columbia Consumer
Video dataset (CCV) \citep{conf/mir/JiangYCEL11} and the TRECVID
Multimedia Event Detection (MED) dataset \citep{over2013trecvid} are
becoming popular. 

\vspace{0.1cm}\noindent\textbf{Feature Representation}\quad Local space-time feature approaches have become the the prevailing strategies due to not requiring non-trivial object tacking and segmentation. In these approaches, local interest points are first detected \citep{laptev2005space} or densely sampled \citep{WangOVS_IJCV15}. Visual descriptors invariant to clutter, appearance and scale are calculated in a spatiotemporal volume formed by the interest points. Different visual descriptors have been proposed to capture the texture, shape and motion information, including 3D-SIFT \citep{scovanner20073}, HOG3D \citep{klaser2008spatio} and local trinary patterns \citep{yeffet2009local}. Among these, dense trajectory features with HOG, HOF and MBH descriptors \citep{wang2013dense} and its variant improved trajectory features \citep{WangOVS_IJCV15}  produce state-of-the-art performance on action recognition. Therefore, we choose improved trajectory feature (ITF) for our low-level feature representation.

\subsection{Zero-Shot Learning}
Zero-shot learning aims to achieve dynamic construction of classifiers for novel classes at testing time based on semantic descriptors provided by humans or existing knowledge bases, rather than labeled examples. This approach  was popularised by the early studies \citep{larochelle2008zerodata,palatucci2009zero_shot,Lampert2009}. Since then numerous studies have been motivated to investigate ZSL due to  the scalability barrier of exhaustive annotation for supervised learning, and the desire to emulate the human ability to learn \emph{from description} with few or no examples. 

\vspace{0.1cm}\noindent\textbf{ZSL Architectures}\quad Various
architectures have been proposed for zero-shot recognition of classes
$\vect{y}$ given data $\matr{X}$. Sequential architectures
\citep{Lampert2009,Fu2014a,Fu2015,Liu2011,zhao2013rtm_groupActivity,lazaridou2014crossModalDSM,norouzi2014_ICLR}
setup classifier/regressor mappings $\matr{Z}=f(\matr{X})$ to predict semantic
representations $\matr{Z}$, followed by a recognition function
$\vect{y}=r(\matr{Z})$. The visual feature mapping $f(\cdot)$ is learned from training data and assumed
to generalise, and the recogniser is given by the human or
external knowledge. Converging architectures
\citep{akata2015outputEmbedding,yang2015mdmtl,paredes2015ez_zsl,Frome2013} setup
energy functions $E(\matr{X},\matr{Z})$ which are positive when $\matr{X}$ and $\matr{Z}$ are from
matching classes and negative otherwise. {In this work, we adopt a
sequential regression approach for simplicity and efficiency of 
closed-form solution, 
and amenability to exploiting the unlabelled data manifold.}

\vspace{0.1cm}\noindent\textbf{Attribute Embeddings}\quad
The most popular intermediate representation for ZSL has been
attributes, where categories are specified in terms of a vector of
binary \citep{Lampert2009,Liu2011,zhao2013rtm_groupActivity} or
continuous \citep{Fu2014a,akata2015outputEmbedding,paredes2015ez_zsl}
attributes. However, this approach suffers inherently from the need to agree
upon a universal attribute ontology, and the scalability barrier of
manually defining each new class in terms of an attribute ontology
that grows with breadth of classes considered
\citep{Fu2014a,akata2015outputEmbedding}. 

\vspace{0.1cm}\noindent\textbf{Word-Vector Embeddings}\quad
While other representations including taxonomic
\citep{akata2015outputEmbedding}, co-occurence
\citep{GanLYZH_AAAI_15,mensink2014costa,Habibian2014} and template-based
\citep{larochelle2008zerodata} have been considered, word-vector space
ZSL
{\citep{Fu2015,akata2015outputEmbedding,Xu_SES_ICIP15,lazaridou2014crossModalDSM,norouzi2014_ICLR,Frome2013}
has emerged as the most effective unsupervised alternative to
attributes. In this approach, the semantic class descriptor $\matr{Z}$ is
generated automatically from existing unstructured text knowledge
bases such as the Wikipedia. In practice, this often means the target $\matr{Z}$
of mapping $\matr{Z}=f(\matr{X})$ is given by the internal representation of a text
modelling neural network \citep{Mikolov2013}. This can be more
intuitively understood as encoding each class name in terms of a
vector describing its co-occurance frequency with other terms in a
text corpus \citep{lazaridou2014crossModalDSM}. In sequential architectures 
the final recognition is typically performed with nearest neighbour (NN) matching
 of the predicted class descriptor \citep{Xu_SES_ICIP15,lazaridou2014crossModalDSM,norouzi2014_ICLR}.}


\vspace{0.1cm}\noindent\textbf{Domain-Shift}\quad
Every ZSL method suffers from the issue of domain shift between the
training class on which the mapping $f(\cdot)$ or energy function
$E(\cdot,\cdot)$ is trained, and the disjoint set of testing classes to
which it is tested on. Although this is a major reason why it is hard
to obtain competitive results with ZSL strategies, it is only recently
this problem has been studied explicitly
\citep{dinu2015improving,Fu2015,paredes2015ez_zsl}. In this work, we
focus primarily on how to mitigate this domain-shift problem in ZSL
for action recognition. That is, by making the training data more
representative thus learning a more general visual feature to semantic space mapping (dataset
augmentation), transductively exploiting both labelled and unlabelled data manifold to
learn an embedding mapping that generalises better to the testing data (manifold regularized
regression), and post-processing corrections to adapt (self-training)
the classifier at the testing time therefore to improve its robustness (hubness
correction) to domain shift. While transductive
\citep{dinu2015improving,Fu2015,Xu_SES_ICIP15} strategies have been
exploited before as post-processing, this is the first time it have
been exploited for learning the embedding itself via manifold
regression. 

\vspace{0.1cm}\noindent\textbf{ZSL Insights}\quad
{Previous studies have provided particular insight into the
  ZSL problem, including \citet{Rohrbach2010,akata2015outputEmbedding}
  who focus on exploring and comparing different class-label
  embeddings (we use word-vectors), \citet{Rohrbach2011} who explores
  scalability to large scale settings, and \citet{dinu2015improving}
  who discusses why ZSL is harder than supervised
  learning due to the hubness problem. Our insights aim to complement
  the above studies by exploring when positive transfer occurs, and
  showing how it is possible to predict this in advance.} 


\subsection{ZSL for Action Recognition}
Despite clear appeal from ZSL, few studies have considered it for action
recognition. Early attribute-centric studies took latent SVM
\citep{Liu2011} and topic model
\citep{Fu2014a,zhao2013rtm_groupActivity} approaches, neither of which
are very scalable for large video datasets. Thus more recent studies
have started to consider unsupervised embeddings including semantic
relatedness \citep{GanLYZH_AAAI_15} and word-vectors
\citep{Xu_SES_ICIP15}. However, most prior ZSL action recognition
studies do not evaluate against a wide range of realistic set of contemporary action
recognition benchmarks, restricting themselves to a single dataset of USAA
\citep{Fu2014a,zhao2013rtm_groupActivity}, or Olympic Sports
\citep{Liu2011}. In this work, we fully explore word-vector-based zero-shot action
 recognition, and demonstrate its superiority to attribute-based
 approaches, despite the latter's supervised ontology construction. Another line of work towards zero-shot action recognition have been studied by \citet{DBLP:conf/iccv/JainGMS15} who proposed to exploit the vast object annotations, images and textual descriptions, e.g. ImageNet \citep{DBLP:conf/cvpr/DengDSLL009}.
{
\subsection{ZSL for Event Detection}
In contrast to action recognition, another line of work on the related task of event detection typically deals with temporally longer multimedia videos. 
The most widely studied test is the TRECVID Multimedia Event Detection
(MED) benchmark \citep{over2013trecvid}. In the zero-shot MED task
(MED 0EK), 20 events are to be detected among a 27K video {(Test Set
  MED)} with no positive examples of each test event available for
training. } 
{
Existing studies \citep{Wu2014,chen2014event,habibian2014composite} typically
discover a `concept space' by extracting frequent terms with pruning
in video metadata (per-video text description) and learning concept
classifiers on the 10K video {Research Set}. Then for each of the 20
events to be detected, a query is generated as a concept vector from
the metadata of the event (textual description of event)
\citep{Wu2014} or an event classifier is learned on 10 positive
examples of the testing event \citep{habibian2014composite}. The
testing videos are finally tested against the concept classifiers and
then matched to the query as inner product between concept detection scores and
query concepts \citep{Wu2014} or through the event classifier
\citep{habibian2014composite}. {Alternatively, visual concept can be mined from noisy online image repositories. In the concept space, a query is then generated from event name and keywords which  are extracted from event definitions \citep{chen2014event}.} These approaches rely on two
assumptions: (1) A large concept training pool (10K video) with
per-video textual description annotated by experts. (2) A detailed
description of the event to be detected is needed to generate the
query. For example a typical event description includes the name -
`Birthday Party', Explication - `A birthday in this context is the
anniversary of a person's birth etc', Object/People - `Decorations,
birthday cake, candles, gifts, etc'. Since detailed per-video
annotations and detailed descriptions of event types are not widely
available in other video databases, in this work we focus on exploring the TRECVID
task with the more challenging but also more broadly applicable setting of
\emph{event name}-driven training and queries only.} {This setting is rarely studied for TRECVID, except in the recent study \citep{DBLP:conf/iccv/JainGMS15} which explores using a Fisher vector to encode compound event/action names.} 

\section{Methodology}

To formalise the problem a list of notations are first given in Table~\ref{tab:Notations}. We have a training video set
$\matr{T}_{tr}=\{\matr{X}_{tr},\vect{y}_{tr}\}$ where $\matr{X}_{tr}=\{\vect{x}_{i}\}_{i=1\cdots n_{l}}$ is the
set of $d_x$ dimensional low-level features, e.g. Fisher Vector
encoded MBH and HOG. For each of the $n_{l}$ labelled training videos 
$\scal{y}_{i}$ is the class names/labels of
each instance, e.g. ``brush hair" and ``handwalk". We also have a set
of testing videos $\matr{T}_{te}=\{\matr{X}_{te},\vect{y}_{te}\}$ with $n_u$ unlabelled testing video
instances. The goal of ZSL is to learn to recognise videos in $\matr{X}_{te}$
whose classes $\vect{y}_{te}$ are disjoint from any seen data at training time:
$\vect{y}_{tr}\cap \vect{y}_{te}=\emptyset$.

\begin{table}[htb]
\centering
\caption{Basic notations.}
\label{tab:Notations}
\resizebox{0.49\textwidth}{!}{
\begin{tabular}{p{2.2cm} p{7cm}}
\hline
\textbf{Notation}                   & \textbf{Description}                     \\ \hline
$\matr{X}\in \mathbb{R}^{d_x\times N}$; $\vect{x}_i$     & Visual feature matrix for N instances; Column representing the $i$-th instance   \\
$\vect{y}\in \mathbb{Z}^{1\times N}$; $\scal{y}_i$      & Integer class labels for N instances; Scalar representing the $i$-th instance      \\
$\matr{Z}\in \mathbb{R}^{d_z\times N}$; $\vect{z}_i$     & Semantic embedding for N instances; Column representing the $i$-th instance       \\
$\matr{K}\in \mathbb{R}^{N\times N}$       & Kernel matrix                            \\
$\matr{A}\in \mathbb{R}^{d_z\times N}$     & Regression coefficient matrix            \\
$f: \matr{X} \to \matr{Z}$                          & Visual to semantic mapping function      \\
$g: \vect{y} \to \matr{Z}$                          & Class name embedding function            \\
$\lambda_A\in \mathbb{R}$							& Ridge regression regularizor			   \\
$\lambda_I\in \mathbb{R}$							& Manifold regression regularizor			\\
$N^{G}_K\in \mathbb{Z}^+$							& KNN Graph parameter for manifold regularizor\\
$N^{st}_K\in \mathbb{Z}^+$							& KNN parameter for Self-Training procedure\\
 \hline
\end{tabular}}
\end{table}

\subsection{Semantic Embedding Space}\label{subsec:semanticembedding}

To bridge the gap between disjoint training and testing classes, we
establish a semantic embedding space $\matr{Z}$ based on word-vectors. In
particular we use a neural network \citep{Mikolov2013} trained on a
100 billion word corpus to realise a mapping $g: \vect{y} \to \matr{Z}$ that
produces a unique $d_z$ dimensional encoding vector of each dictionary
word.  

\vspace{0.1cm}\noindent\textbf{Compound Names}\quad The above
procedure only deals with class names that are unigram dictionary
words. To process compound names  commonly occurring in action
datasets, e.g. ``brush hair" or ``ride horse", that do not exist as
individual tokens in the corpus, we exploit compositionally of the
semantic space \citep{mitchell2008}. Various composition methods have
been proposed \citep{mitchell2008,milajevs2014sentenceComposition}
including additive, multiplicative and others, but our experiments
showed no significant to using others besides addition, so we stick
with simple additive composition.  

Suppose the $ith$ class name $\scal{y}_i$ is composed of words $\{y_{ij}\}_{j=1\cdots w}$. We generate a single $d_z$ dimensional vector $\vect{z}$ out of the word-vector $\scal{y}_{i}$ by a averaging word-vectors for constituent words $\{y_{ij}\}$:
\begin{equation}\label{eq:AdditiveModel}
\vect{z}_i=\frac{1}{w}\cdot\sum_{j=1}^{w}g(y_{ij})
\end{equation}

\subsection{Visual to Semantic Mapping}\label{sec:embedLearn}
\noindent\textbf{Mapping by Regression:}\quad
In order to map video features into the semantic embedding space constructed above, we train a regression model $f: X \to Z$ from $d_x $ dimensional low-level visual feature space to the $d_z$ dimensional embedding space. 
The regression is trained using training instances $\matr{X}_{tr}=\{\vect{x}_i\}_{i=1\cdots n_{l}}$ and the corresponding embedding $\matr{Z}_{tr}=g(\vect{y}_{tr})$ of the instance class name $\scal{y}$ as the target value. 
Various methods have previously been used for this task including linear support vector regression (SVR) \citep{Fu2014a,Fu2015,Xu_SES_ICIP15} and more complex multi-layer neural networks \citep{Socher2013,lazaridou2014crossModalDSM,yang2015mdmtl}.
Since we will use fisher vector encoding \citep{perronnin2010improving} for features $\matr{X}$, we can easily apply simple linear regression for $f(\cdot)$. Specifically, we use $l_2$ regularized linear regression (aka ridge regression) to learn the visual to semantic mapping. 

\noindent\textbf{Kernel Ridge Regression:}\quad
The fisher vector encoding generates a very high dimensional feature
$2\times d_{descr} \times N_k$ where $N_k$ is the number of components
in the Gaussian Mixture Model (GMM) and $d_{descr}$ is the dimension
of raw descriptors. This usually results in many more feature
dimensions than training samples. Thus we use the representer theorem
\citep{scholkopf2002lwk} and formulate a kernelized ridge regression
with a linear kernel in Eq~(\ref{eq:LinearKernel}). {The
  benefit of kernelised regression is to reduce computation as the
  closed-form solution to $\matr{A}$ only involves computing the inverse of a
  $N\times N$  rather than a $d_x \times d_x$ matrix where $N< d_x$.} 

\begin{equation}
\scal{k}(x_i,x_j)=\sum\limits_{d=1}^{d_x}(x_{id}\cdot x_{jd})\label{eq:LinearKernel}
\end{equation}

The visual features $\vect{x}$ can be then projected into semantic space via Eq~(\ref{eq:Visual2SemanticMap}) where $\vect{a}_{j}$ is the $j$th column of regression parameter matrix $\matr{A}$.

\begin{equation}
f(\vect{x}) =\sum\limits_{j=1}^{n_{l}}  \vect{a}_{j}\scal{k}(\vect{x},\vect{x}_j) \label{eq:Visual2SemanticMap}
\end{equation}

To improve the generalisation of the regressor, we add the $l_2$ regularizer $||f||_{\matr{K}}^2= Tr(\matr{A} \matr{K} \matr{A}^T)$ to reduce overfitting by penalising extreme values in the regression matrix. This gives the kernel ridge regression loss:

\begin{equation}\label{eq:RidgeRegress}
\begin{split}
&\min_f\frac{1}{n_{l}}\sum\limits_{i=1}^{n_{l}}||\vect{z}_i-f(\vect{x}_i)||^2_2 + \gamma ||f||^2_{K}\\
&\min_\matr{A}\frac{1}{n_{l}}Tr\left((\matr{Z}-\matr{AK})^T(\matr{Z}-\matr{AK})\right)+\gamma Tr(\matr{AKA}^T)
\end{split}
\end{equation}

\noindent where the regression targets are generated by the vector representation of each class name $\vect{z}_i=g(\scal{y}_i)$ and $\matr{Z}=[\vect{z}_1 \: \vect{z}_2 \cdots]_{d_z\times n_l}$, $\matr{A}$ is the $d_z  \times n_l$ regression coefficient matrix, $\matr{K}$ is the $n_l\times n_l$ kernel matrix and {$n_l$ is the number of labelled training instances}. The loss function is convex with respect to the $\matr{A}$.
Taking derivatives w.r.t $\matr{A}$ and setting the gradient to 0 leads to the following closed-form solution where $\mathbf{I}$ is the identity matrix.

\begin{equation}\label{eq:ManifoldCloseform}
\matr{A}={\matr{Z}}\left(\matr{K}+\gamma_A n_{l} \mathbf{I}\right)^{-1}
\end{equation} 

The above mapping by Kernel Ridge Regression provides a simple solution to embed visual instances into semantic space. However the simple ridge regression only considers limited labelled training data $\matr{X}_{tr}$ without exploiting the underline structure of the manifold on both labelled and unlabelled data nor any additional related labelled data from other datasets. In the following sections, we introduce two approaches to improve the quality of mapping: (1) \textit{Manifold-Regularized Regression} and (2) \textit{Data Augmentation}.

\subsubsection{Manifold Regularized Regression}\quad
As discussed earlier, conventional regularization provides poor ZSL
due to disjoint training and testing classes. To  improve recognition
of testing classes, we explore transductive semi-supervised
regression. The idea is to exploit unlabelled testing data $\matr{X}_{te}$ to
discover the manifold structure in the zero-shot classes,
and preserve this structure in the semantic space after
visual-semantic mapping. Therefore, this is also known as manifold regularization. {Note that we use \textit{labelled} to refer to training data $\matr{X}_{tr}$ and \textit{unlabelled} to refer to testing data $\matr{X}_{te}$. So we use semi-supervised manifold regularization in a transductive way, requiring access to the unlabelled/testing data $\matr{X}_{te}$ during the training phase.}

To that end, we introduce manifold laplacian regularization
\citep{belkin2006manifold} into the ridge regression formulation. This
additional regularization term ensures that if two videos are close to
each other in the visual feature space, this relationship should be
kept in the semantic space as well.  

We model the manifold by constructing a symmetric K nearest neighbour (KNN)
graph $\matr{W}$ on the all  $n_{l}+n_{u}$ instances where $n_l=|\matr{T}_{tr}|$ denotes the
number of labelled training instances and $n_u=|\matr{T}_{te}|$ denotes the number of
unlabelled testing instances. {The KNN Graph is constructed by
  first computing a linear kernel matrix between all instances. Then
  for each instance we select the top K neighbours and assign an edge
  between these nodes. This gives us a directed graph which is then
  symmetrized by converting to an undirected graph by connecting nodes
  with any directed edge between them.} Let $\matr{D}$ be a diagonal matrix
with 
$d_{ii}=\sum_{j=1}^{n_l+n_u} w_{ij}$, we get the graph laplacian
matrix $\matr{L}=\matr{D}-\matr{W}$. The manifold regularizer is then written as:

\begin{equation}\label{eq:ManifoldRegularizor}
\resizebox{0.87\columnwidth}{!}{
$
\begin{split}
||f||_I^2 & = \frac{1}{2}\sum\limits_{i,j}^{n_l+n_u} w_{ij} ||f(\vect{x}_i)-f(\vect{x}_j)||^2_2\\
& = \frac{1}{2}\sum\limits_{i,j}w_{ij} f^\top(\vect{x}_i)f(\vect{x}_i)+\frac{1}{2} \sum\limits_{i,j}w_{ij} f^\top(\vect{x}_j)f(\vect{x}_j) \\
& \quad\quad - \sum\limits_{i,j}w_{ij} f^\top(\vect{x}_i)f(\vect{x}_j)\\
& = \sum\limits_{i} d_{ii} f^\top(\vect{x}_i)f(\vect{x}_i) - \sum\limits_{i,j}w_{ij} f^\top(\vect{x}_i)f(\vect{x}_j)\\
\end{split}$
}
\end{equation}

Further denoting $\mathbf{f}=[f(\vect{x}_1)\; f(\vect{x}_2) \cdots \;f(x_{n_l+n_u})]=\matr{A}\matr{K}$. Eq.~(\ref{eq:ManifoldRegularizor}) can be rewritten in matrix form as:

\begin{equation}
\begin{split}
||f||_I^2 & = Tr(\mathbf{f}^\top \mathbf{f}\matr{D}) - Tr(\mathbf{f}^\top \mathbf{f}\matr{W} )\\
& = Tr(\mathbf{f}^\top \mathbf{f}\matr{L} )\\
& = Tr(\matr{K}^\top\matr{A}^\top\matr{A}\matr{K}\matr{L})
\end{split}
\end{equation}

\noindent where $\matr{K}$ is a $(n_l+n_u) \times (n_{l}\times n_u)$ dimensional kernel matrix constructed upon all labelled and unlabelled instances via Eq~(\ref{eq:LinearKernel}). Combining all regularization terms we obtain the overall loss function in Eq~(\ref{eq:ManifoldRegressionLossFcn}), where for simplicity we denote  $\matr{J}=\left[\begin{array}{ccc}
\mathbf{I}_{n_l\times n_l} & \mathbf{0}_{n_l\times n_u}\\ \mathbf{0}_{n_u\times n_l} & \mathbf{0}_{n_u\times n_u}\end{array}\right]$ and $\tilde{\matr{Z}}=[\matr{Z}_{tr}\quad \mathbf{0}_{d_z \times n_u}]$. The final loss function can be thus written in the matrix form as:

{
\begin{equation}\label{eq:ManifoldRegressionLossFcn}
\resizebox{0.95\columnwidth}{!}{
$
\begin{split}
&\min_\matr{A}\frac{1}{n_{l}}Tr\left((\tilde{\matr{Z}}-\matr{AKJ})^\top(\tilde{\matr{Z}}-\matr{AKJ})\right) + \gamma_A Tr(\matr{A} \matr{K} \matr{A}^\top)  \\
&\quad\quad + \frac{\gamma_I}{(n_l+n_u)^2} Tr( \matr{K}^\top\matr{A}^\top\matr{AKL})
\end{split}
$
}
\end{equation}}

\noindent 
The loss function is convex w.r.t. the  $ d_z \times (n_l+n_u)$ regression coefficient matrix $\matr{A}$. 
A closed-form solution to $\matr{A}$ can be obtained in the same way as {Kernel Ridge Regression}.

\begin{equation}\label{eq:ManifoldCloseform}
\matr{A}=\tilde{\matr{Z}}\left(\matr{KJ}+\gamma_A n_{l} \mathbf{I} + \frac{\gamma_I n_l}{(n_l+n_u)^2}\matr{KL}\right)^{-1}
\end{equation} 

\noindent Eq~(\ref{eq:ManifoldCloseform}) provides an efficient way to learn the visual to semantic mapping due to the closed-form solution compared to alternative iterative approaches \citep{Fu2014a,Habibian2014}. At testing time, the mapping can be efficiently applied to project new videos into the embedding with Eq.~(\ref{eq:Visual2SemanticMap}). Note when $\gamma_I=0$ manifold regression becomes exactly kernel regression.

\subsubsection{Improving the Embedding with Data Augmentation}\label{sec:augment}
As discussed, the mapping often generalises poorly because: (i) actions are visually complex and ambiguous, and (ii) even a mapping well learned for training categories may not generalise well to testing categories as required by ZSL, because the volume of training data is small compared to the complexity of a general visual to semantic space mapping. The manifold regression described previously ameliorates the latter issues, but we next discuss a complementary strategy of data augmentation. 

Another way to further mitigate both of these problems is by
augmentation with any available auxiliary dataset which need not
contain classes in common with the target testing dataset $\matr{T}^{trg}_{te}$in which zero-shot recognition is performed. This will
provide more data to learn a better generalising regressor
$\vect{z}=f(\vect{x})$. We formalize the data augmentation problem as follows. We
denote the target dataset as $\matr{T}^{trg}=\{\matr{X}^{trg},\vect{y}^{trg}\}$ split into training set $\matr{T}^{trg}_{tr}=\{\matr{X}^{trg}_{tr},\vect{y}^{trg}_{tr}\}$ and zero-shot testing set $\matr{T}^{trg}_{te}=\{\matr{X}^{trg}_{te},\vect{y}^{trg}_{te}\}$. Zero-shot recognition is performed on the testing set of the target dataset (e.g. HMDB51). There are
$n_{aux}$ other available auxiliary datasets $\matr{T}^{aux}_{i=1\cdots
  n_{aux}}=\{\matr{X}^{aux}_i,\vect{y}^{aux}_i\}$ (e.g. UCF101, Olympic Sports and CCV). We propose to improve the regression by merging the target dataset
training data and all auxiliary sets. The auxiliary
dataset class names $\vect{y}^{aux}_i$ are projected into the embedding space
with  $\matr{Z}^{aux}_i=g(\vect{y}^{aux}_i)$. {The auxiliary instances $\matr{X}^{aux}$ are aggregated
with the target training data as $\matr{X}_{tr}=[\matr{X}^{trg}_{tr} \; \matr{X}^{aux}_1 \; \cdots \; \matr{X}^{aux}_{n_{aux}}]$ and $\matr{Z}_{tr}=[\matr{Z}^{trg}_{tr} \; \matr{Z}^{aux}_{1}
  \; \cdots \; \matr{Z}^{aux}_{n_{aux}}]$ where $\matr{Z}^{trg}_{tr}=g(\vect{y}^{trg}_{tr})$. The augmented training data $\matr{X}_{tr}$ and class embeddings $\matr{Z}_{tr}$ are used together to train the regressor $f$.}
%

{To formulate the loss function in matrix form we denote
  $n^{trg}_{l}=|\matr{T}^{trg}_{tr}|$, $n^{trg}_{u}=|\matr{T}^{trg}_{te}|$,
  $n^{aux}_l=\sum\limits_i|\matr{T}^{aux}_i|$. Let $\tilde{\matr{K}}$ be the
  $(n^{trg}_{tr}+n^{trg}_{te}+n^{aux}_l) \times
  (n^{trg}_{tr}+n^{trg}_{te}+n^{aux}_l)$ dimensional kernel matrix on
  all target and auxiliary data, and $\tilde{\matr{L}}$ is the corresponding
  graph laplacian. We then write the block structured $\tilde{\matr{J}}$
  matrix as:}

{
\begin{equation}
\tilde{\matr{J}}=
\begin{bmatrix}
\mathbf{I}_{(n_l^{trg}+n_l^{aux})\times (n_l^{trg}+n_l^{aux})} & \mathbf{0}\\ 
\mathbf{0} & \mathbf{0}_{n_u^{trg}\times n_u^{trg}}
\end{bmatrix}
\end{equation} 
}

\noindent{The loss function of manifold regularized regression with data augmentation is thus written in a matrix form as:}
  
{
\begin{equation}\label{eq:DataAugLossFcn}
\resizebox{0.90\columnwidth}{!}{
$
\begin{split}
&  \min_{\matr{A}}\frac{1}{(n^{trg}_{tr}+n^{aux}_l)}Tr\left((\tilde{\matr{Z}}_{tr}-\matr{A}\tilde{\matr{K}}\tilde{\matr{J}})^T(\tilde{\matr{Z}}_{tr}-\matr{A}\tilde{\matr{K}}\tilde{\matr{J}})\right) \\
& \quad +\gamma_A Tr(\matr{A}\tilde{\matr{K}}\matr{A}^T) \\
& \quad+ \frac{\gamma_I}{(n^{trg}_{tr}+n^{trg}_{te}+n^{aux}_l)^2}Tr(\tilde{\matr{K}}^T\matr{A}^T\matr{A}\tilde{\matr{K}}\tilde{\matr{L}})\\
\end{split}
$}
\end{equation}  
}

In the same way as before, we obtain the closed-form solution to $\matr{A}$:

\begin{equation}
\resizebox{0.68\columnwidth}{!}{
$
\begin{split}
&\matr{A}=\tilde{\matr{Z}_{tr}}\bigg(\tilde{\matr{K}}\tilde{\matr{J}} + \gamma_A (n^{trg}_{tr}+n^{aux}_{l})\mathbf{I}+\\
& \quad\quad \frac{\gamma_I (n^{trg}_{tr}+n^{aux}_{l})}{(n^{trg}_{tr}+n^{trg}_{te}+n^{aux}_l)^2}\tilde{\matr{K}}\tilde{\matr{L}}\bigg) ^{-1}
\end{split}
$}
\end{equation}

\noindent where by setting $\gamma_I=0$ we obtain a kernel ridge regression with only data augmentation. {This solution can be conveniently implemented in a single line of Matlab.}



\subsection{Zero-Shot Recognition}

Given the trained mappings $f(\cdot)$ and $g(\cdot)$ we can now complete the zero-shot learning task. To classify a testing instance $\vect{x}^*\in \matr{X}_{te}$, we apply nearest neighbour matching of the projected testing instance $f(\vect{x}^*)$ against the vector representations of all the testing classes $g(\scal{y})$ (named the prototype throughout this paper): 

\begin{equation}\label{eq:zsl_nn}
\hat{\scal{y}} = \arg\min_{\scal{y}\in \vect{y}_{te}}\|f(\vect{x}^*) - g(\scal{y})\|
\end{equation}

\noindent Distances in such embedding spaces have been shown to be best measured using the cosine metric  \citep{Mikolov2013,Fu2014a}. Thus we $l_2$ normalise each data point, making Euclidean distance effectively equivalent to cosine distance  in this space.


\subsubsection{Ameliorating Domain Shift by Post Processing}\label{sec:postprocess}

In the previous two sections we introduced two methods to improve the embedding $f$ for ZSL. In this section we now  discuss two post-processing strategies to further reduce the impact of domain shift.

\noindent\textbf{Self-training for Domain Adaptation}\quad
The domain shift induced by applying $f(\cdot)$ trained on $\matr{X}_{tr}$ to data of different statistics $\matr{X}_{te}$ means the projected data points $f(\matr{X}_{te})$ do not lie neatly around the corresponding class projections/prototypes $g(\vect{y}_{te})$ \citep{Fu2015}.  To ameliorate this domain shift, we explore transductive self-training to adjust unseen class prototypes to be more comparable to the projected data points. For each category prototype $g(\scal{y}^*), \scal{y}^*\in \vect{y}_{te}$ we search for the $N^{st}_K$ nearest neighbours among the unlabelled testing instance projections, and re-define the adapted prototype $\tilde{g}(\scal{y}^*)$ as the average of those $N^{st}_K$ neighbours. Thus if $NN_K(g(\scal{y}^*))$ denotes the set of K nearest neighbours of $g(\scal{y}^*)$, we have:
\begin{equation}\label{eq:self_train}
\tilde{g}(\scal{y}^*) := \frac{1}{N^{st}_K}\sum_{f(\vect{x}^*)\in NN_K(g(\scal{y}^*))}^{N^{st}_K} f(\vect{x}^*)
\end{equation}
The adapted prototypes $\tilde{g}(\scal{y}^*)$ are now more directly comparable with the testing data for matching using Eq.~(\ref{eq:zsl_nn}). 

\noindent\textbf{Hubness Correction}\quad
One practical effect of the ZSL domain shift was elucidated in \citet{dinu2015improving}, and denoted the `Hubness' problem. Specifically, after the domain shift, there are a small set of `hub' test-class prototypes that become nearest or K nearest neighbours to the majority of testing samples in the semantic space, while others are NNs of no testing instances. This results in poor accuracy and highly biased predictions with the majority of testing examples being assigned to a small minority of classes.   We therefore explore the simple solutions proposed by \citet{dinu2015improving} which takes into account the global distribution of zero-shot samples and prototypes. This method is transductive as with self-training and manifold-regression. Specifically, we considered two alternative approaches: \textit{Normalized Nearest Neighbour} (NRM) and \textit{Globally Corrected} (GC).

The NRM approach eliminates the bias towards hub prototypes by normalizing the distance of each prototype to all testing samples prior to performing Nearest Neighbour classification as defined in Eq~(\ref{eq:zsl_nn}). More specifically, denote the distance between prototype $\scal{y}_j$ and testing sample $\{\vect{x}_i^*\}_{i=1\cdots n_u}$ as $d_{ij}=||f(\vect{x}_i^*)-g(\scal{y}_j)||$. We then $l_2$ normalize the distances between prototype $\scal{y}_j$ and all $n_u$ testing samples in Eq~(\ref{eq:NRM_distance}). This normalized distance $\hat{d}_{ij}$ replaces the original distance $d_{ij}$ for doing nearest neighbour matching in Eq.~(\ref{eq:zsl_nn}).

\begin{equation}\label{eq:NRM_distance}
\tilde{d}_{ij}=d_{ij}/\sqrt{\sum\limits_i^{n_u} d^2_{ij}}
\end{equation}

The alternatively GC approach damps the effect of hub prototypes by
using ranks rather than the original distance measures. We denote the
function $Rank(\scal{y} , \vect{x}_i^*)$ as the rank of testing sample $\vect{x}_i^*$ w.r.t
the distance to $\scal{y}$. {Specifically, the rank function is
  defined as Eq~(\ref{eq:RankFcn}) where $\mathbbm{1}$ is the
  indicator function.} 

{
\begin{equation}\label{eq:RankFcn}
\begin{split}
&Rank(\scal{y},\vect{x}_i^*)=\\
&\sum\limits_{\vect{x}_j^*\in \matr{X}_{te}\setminus \vect{x}_i^*}\mathbbm{1}(||f(\vect{x}_j^*)-g(\scal{y})||\leq ||f(\vect{x}_i^*-g(\scal{y}))||)
\end{split}
\end{equation}}

 The rank function always return an integer value between $0$ and $|\matr{X}_{te}|-1$. Thus the label of testing sample $x_i^*$ can be predicted by Eq~(\ref{eq:GlobalCorrectedPredict}) in contrast to simple nearest by neighbour Eq~(\ref{eq:zsl_nn}).

\begin{equation}\label{eq:GlobalCorrectedPredict}
\hat{\scal{y}} = \arg\min_{\scal{y}\in \vect{y}_{te}} Rank(\scal{y},\vect{x}_i^*)
\end{equation}

{Note, both strategies {do not} alter the ranking of testing
  samples w.r.t. each prototype. However, the ranking of prototypes
  w.r.t. each testing sample is altered thus potentially improves the
  quality of NN matching. Overall, due to the nature of a retrieval task
  which depends on the ranking of testing samples w.r.t. prototypes,
  the performance of retrieval task is not affected by the two hubness
  correction methods.} 
 

\subsection{Multi-Shot Learning}\label{sect:MulZeroLearning}

Although our focus is zero-shot learning, we also note that the semantic embedding space provides an alternative representation for conventional supervised learning. For multi-shot learning, we  map all data instances $\matr{X}$ into the semantic space using projection $\matr{Z}=f(\matr{X})$, and then simply train SVM classifiers with linear kernel using the $l_2$ normalised projections $f(\matr{X})$ as data. In the testing phase, testing samples are projected into embedding space via the mapping $f(\matr{X})$ and categorised using the SVM classifiers.

\section{Experiments}
\subsection{Datasets and Settings}
\begin{figure*}[!htb]
\centering
\subfigure[HMDB51]{\includegraphics[width=0.24\linewidth]{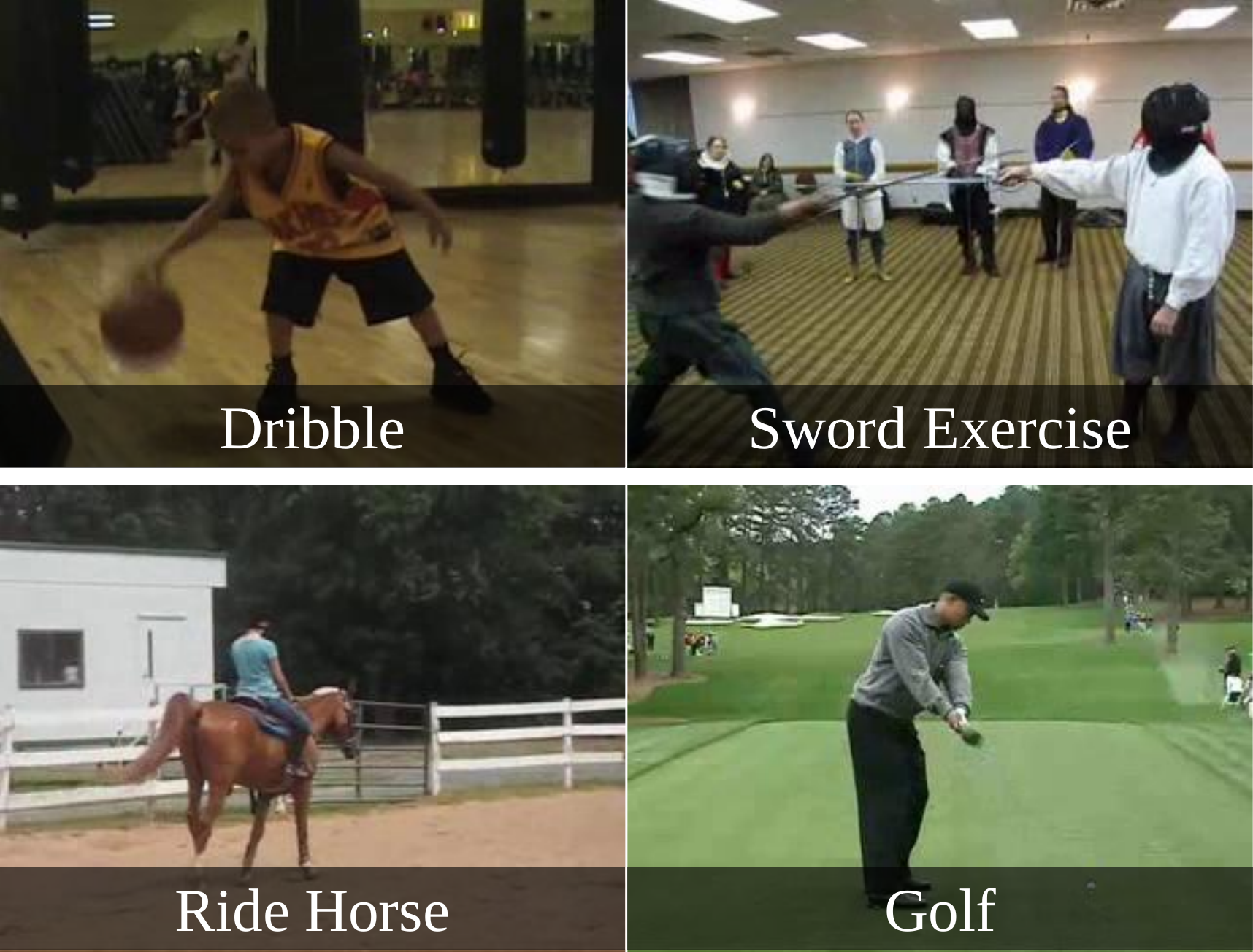}}
\subfigure[UCF101]{\includegraphics[width=0.24\linewidth]{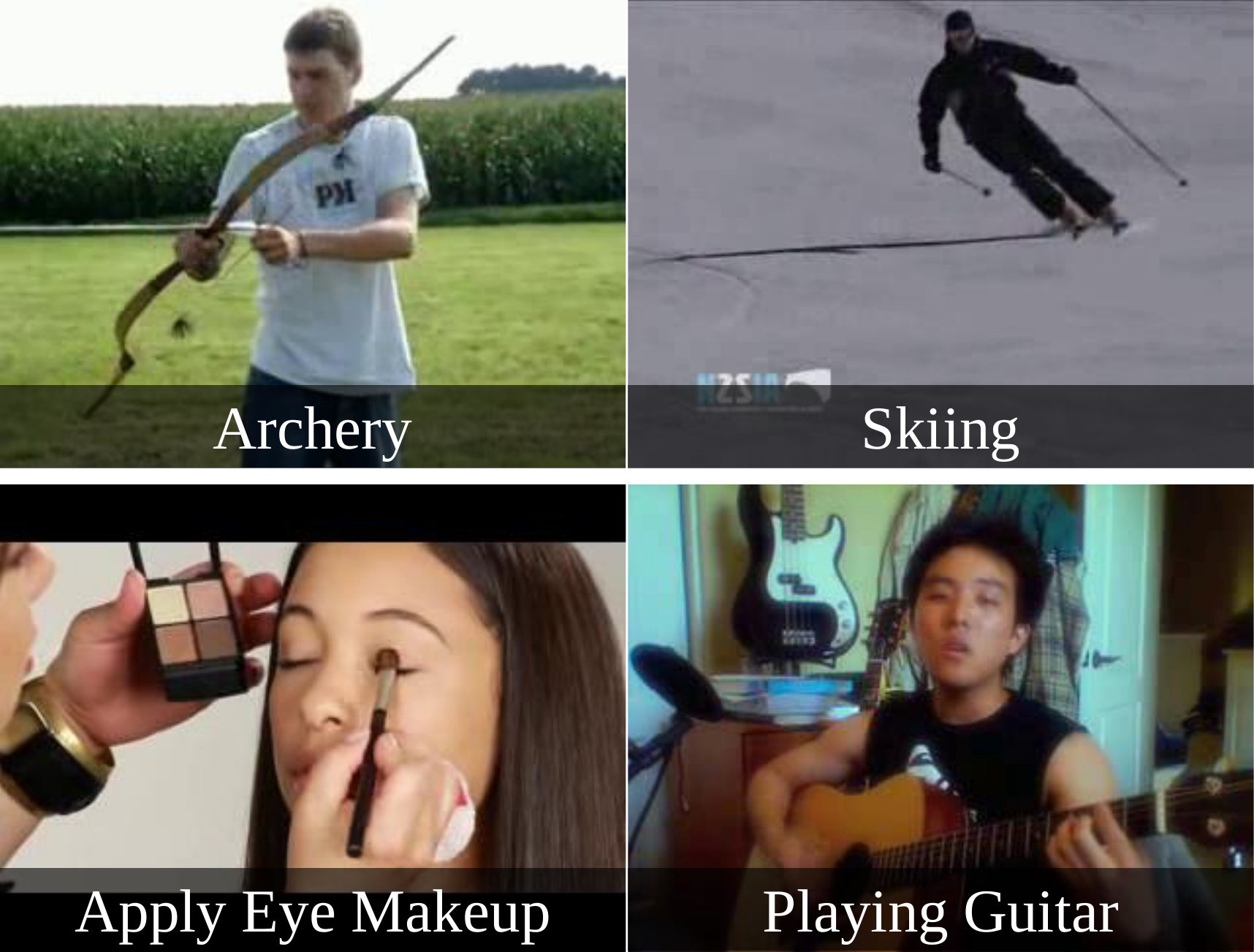}}
\subfigure[OlympicSports]{\includegraphics[width=0.24\linewidth]{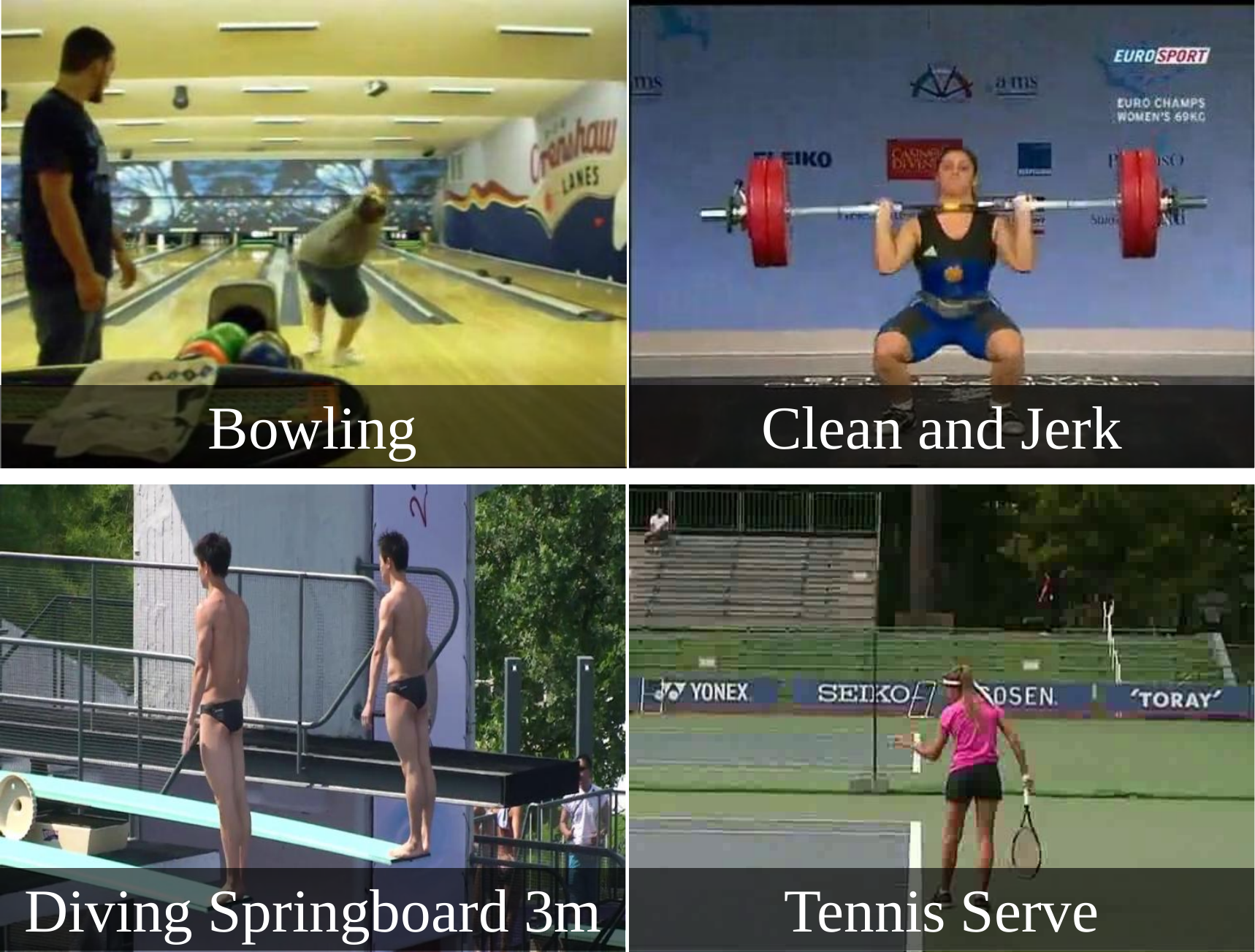}}
\subfigure[$CCV$]{\includegraphics[width=0.24\linewidth]{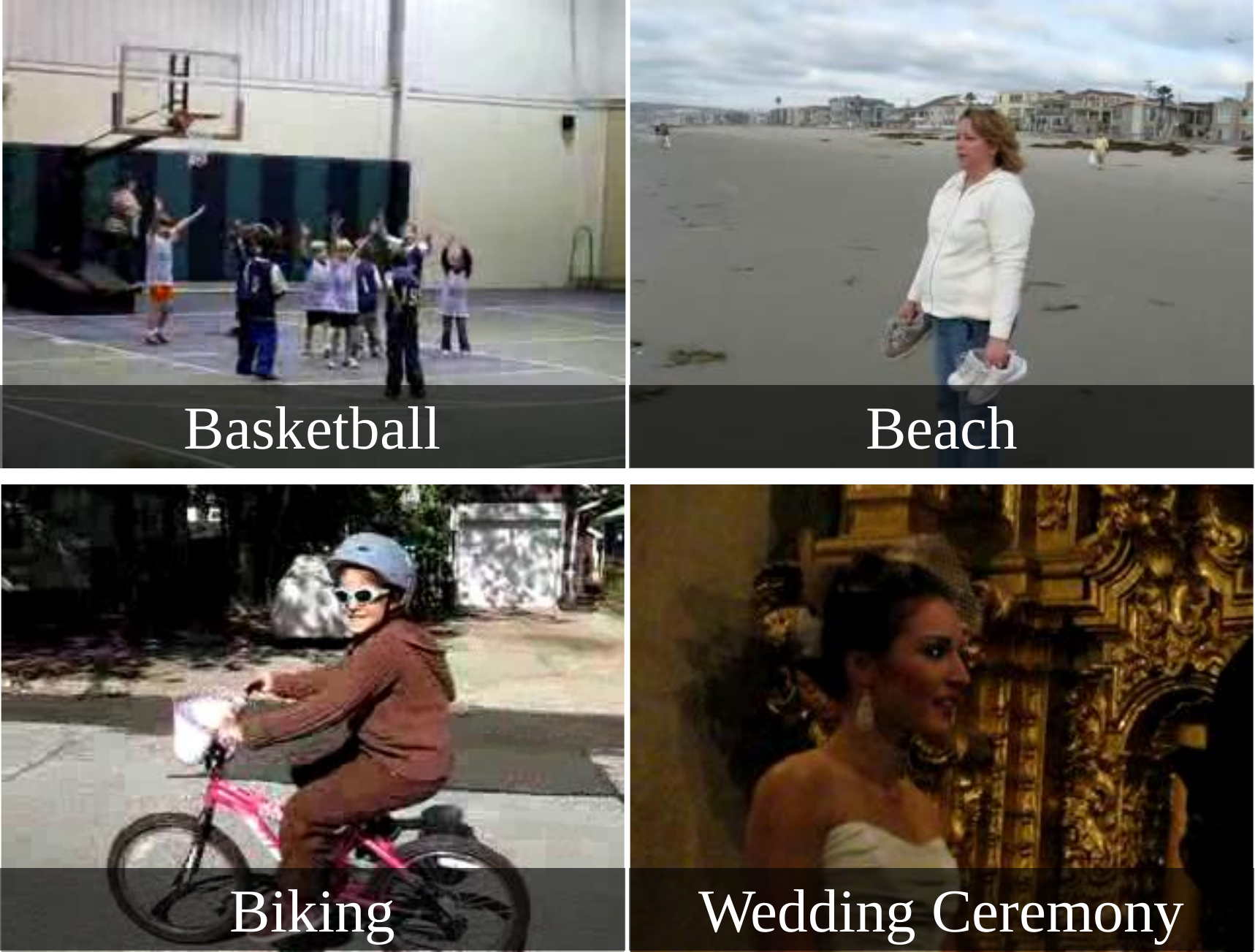}}
\caption{Example frames for different action datasets.}\label{fig:DatasetExampleFrames}
\end{figure*}

\noindent\textbf{Datasets:}\quad Experiments are performed on 5 popular contemporary action recognition and event detection datasets including A Large Human Motion Database (HMDB51)~\citep{Kuehne2011}, UCF101~\citep{Soomro2012}, Olympic Sports~\citep{NieblesCL_eccv10} and Columbia Consumer Video (CCV)~\citep{conf/mir/JiangYCEL11}. HMDB51 is specifically created for human action recognition. It has 6766 videos from various sources with 51 categories of actions. 
UCF101 is an action recognition dataset of 13320 realistic action videos, collected from YouTube, with 101 action categories. Olympic Sports is collected from YouTube, and is mainly focused on sports events. It has 783 videos with 16 categories of events. CCV contains 9682 YouTube videos over 20 semantic categories. We illustrate some example frames in Fig.~\ref{fig:DatasetExampleFrames}.  The action/event category names are presented in Table~\ref{tab:CategoryNames}. We also evaluate USAA~\citep{Fu2014a} -- a subset of CCV specifically annotated with attributes -- in order to facilitate comparison against attribute centric ZSL approaches. 
In addition to above action/event datasets, we also studied a large complex event dataset - TRECVID MED 2013. There are five components to the dataset including Event Kit training, Background training, test set MED, test set Kindred and Research Set.  \textcolor{black}{We use standard test set MED for zero-shot testing data and Event Kit  as training data}.

\begin{table*}[!th]
\centering
\caption{Category names of each dataset.}
\resizebox{0.99\textwidth}{!}{
\begin{tabular}{p{2cm}|p{18cm}}
\toprule
\textbf{Dataset}        & \multicolumn{1}{c}{\textbf{Category Names}}                                                                                                                                                                                                                                                                                                                                                                                                                                                                                                                                                                                                                                                                                                                                                                                                                                                                                                                                                                                                                                                                                                                                                                                                                                                                                                                                                                                                                            \\ \hline
HMDB51         & brush\_hair, cartwheel, catch, chew, clap, climb, climb\_stairs, dive, draw\_sword, dribble, drink, eat, fall\_floor, fencing, flic\_flac, golf, handstand, hit, hug, jump, kick, kick\_ball, kiss, laugh, pick, pour, pullup, punch, push, pushup, ride\_bike, ride\_horse, run, shake\_hands, shoot\_ball, shoot\_bow, shoot\_gun, sit, situp, smile, smoke, somersault, stand, swing\_baseball, sword, sword\_exercise, talk, throw, turn, walk, wave                                                                                                                                                                                                                                                                                                                                                                                                                                                                                                                                                                                                                                                                                                                                                                                                                                                                                                                                                                                                  \\ \hline
UCF101         & Apply Eye Makeup, Apply Lipstick, Archery, Baby Crawling, Balance Beam, Band Marching, Baseball Pitch, Basketball Shooting, Basketball Dunk, Bench Press, Biking, Billiards Shot, Blow Dry Hair, Blowing Candles, Body Weight Squats, Bowling, Boxing Punching Bag, Boxing Speed Bag, Breaststroke, Brushing Teeth, Clean and Jerk, Cliff Diving, Cricket Bowling, Cricket Shot, Cutting In Kitchen, Diving, Drumming, Fencing, Field Hockey Penalty, Floor Gymnastics, Frisbee Catch, Front Crawl, Golf Swing, Haircut, Hammer Throw, Hammering, Handstand Pushups, Handstand Walking, Head Massage, High Jump, Horse Race, Horse Riding, Hula Hoop, Ice Dancing, Javelin Throw, Juggling Balls, Jump Rope, Jumping Jack, Kayaking, Knitting, Long Jump, Lunges, Military Parade, Mixing Batter, Mopping Floor, Nun chucks, Parallel Bars, Pizza Tossing, Playing Guitar, Playing Piano, Playing Tabla, Playing Violin, Playing Cello, Playing Daf, Playing Dhol, Playing Flute, Playing Sitar, Pole Vault, Pommel Horse, Pull Ups, Punch, Push Ups, Rafting, Rock Climbing Indoor, Rope Climbing, Rowing, Salsa Spins, Shaving Beard, Shotput, Skate Boarding, Skiing, Skijet, Sky Diving, Soccer Juggling, Soccer Penalty, Still Rings, Sumo Wrestling, Surfing, Swing, Table Tennis Shot, Tai Chi, Tennis Swing, Throw Discus, Trampoline Jumping, Typing, Uneven Bars, Volleyball Spiking, Walking with a dog, Wall Pushups, Writing On Board, Yo Yo \\ \hline
Olympic Sports & basketball layup, bowling, clean and jerk, discus throw, hammer throw, high jump, javelin throw, long jump, diving platform 10m, pole vault, shot put, snatch, diving springboard 3m, tennis serve, triple jump, vault                                                                                                                                                                                                                                                                                                                                                                                                                                                                                                                                                                                                                                                                                                                                                                                                                                                                                                                                                                                                                                                                                                                                                                                                                                    \\ \hline
CCV            & Basketball, Baseball, Soccer, IceSkating, Skiing, Swimming, Biking, Cat, Dog, Bird, Graduation, Birthday, WeddingReception, WeddingCeremony, WeddingDance, MusicPerformance, NonmusicPerformance, Parade, Beach, Playground                                                                                                                                                                                                                                                                                                                                                                                                                                                                                                                                                                                                                                                                                                                                                                                                                                                                                                                                                                                                                                                                                                                                                                                                                               \\ \bottomrule
\end{tabular}\label{tab:CategoryNames}}
\end{table*}

\vspace{0.1cm}\noindent\textbf{Visual Feature Encoding:}\quad For each video we extract improved trajectory feature (ITF) descriptors  \citep{Wang2013} and encode them with Fisher Vectors (FV). We first compute ITF with 3 descriptors (HOG, HOF and MBH). We apply PCA to reduce the dimension of descriptors by half which results in descriptors with $198$ dimensions in total. Then we randomly sample 256,000 descriptors from each of the 5 action/event datasets and learn a Gaussian Mixture Model with $128$ components from the combined training descriptors. Finally the dimension of FV encoded feature is equal to $d_x=2\times 128 \times 198=50688$. The visual feature for TRECVID MED 2013 dataset was extracted using ITF with HOG and MBH descriptors encoded with Fisher Vectors.  We use the FV encoded feature provided by \citet{Habibian2014}.

\vspace{0.1cm}\noindent\textbf{Semantic Embedding Space:}\quad  We adopted the skip-gram neural network model \citep{Mikolov2013} trained on the Google News dataset (about 100 billion words). This neural network can then encode any of approximately 3 million unique worlds as a $d_z=300$ dimension vector. 


\subsection{Zero-shot Learning on Actions and Events}\label{sect:exp_ZSL}

\noindent\textbf{Data Split:}\quad Because there is no existing zero-shot learning evaluation protocol for most existing action and event datasets we propose our own splits\footnote{The data split will be released on our website}. We first propose a 50/50 category split for all datasets. Visual to semantic space mappings are trained on the $50\%$ training categories, and the other $50\%$ are held out unseen for testing time. We randomly generate 50 independent splits and take the mean accuracy and standard deviation for  evaluation. Among the 50 splits, all categories are evaluated as testing classes, and the frequency is evenly distributed.

\subsubsection{Evaluation of Components} 
To evaluate the efficacy of each component we considered an extensive combination of blocks including manifold regularizer, self-training, hubness correction and data augmentation. Specifically we evaluated the following options for each component.
\begin{itemize}
\item \textbf{Data Augmentation:} Using only within target dataset
  training data (\text{\sffamily X}) to learn the embedding $f(\vect{x})$, or
  also borrowing data from the auxiliary datasets
  (\checkmark). (Section~\ref{sec:augment}). For each of the four
  datasets HMDB51, UCF101, Olympic Sports and CCV, the other three
  datasets are treated as the auxiliary sets. {{Note, there
      are overlapping categories between the auxiliary and target  sets in
      the sense of exact name match. For instance, the action class
      \textit{Biking} exists in both UCF101 and CCV. To avoid
      violating the zero-shot assumption we exclude these exact
      matching classes in the auxiliary set. However, we consider that
      semantic overlaps, e.g. \textit{Biking} in UCF101 and
      \textit{Ride Bike} in HMDB51, should not be excluded because
      recognizing such paraphrase of action category is the problem to
      be solved by zero-shot learning and exploiting such semantic
      relatedness is unique to word-vector embedding approach.}}
\item \textbf{Embedding:} We compare ridge regression (RR) with manifold regularized ridge regression (MR) (Section~\ref{sec:embedLearn}). 
\item \textbf{Self Training:} With (\checkmark) or without (\text{\sffamily X}) self-training before matching (Section~\ref{sec:postprocess}).
\item \textbf{Matching Strategy:} We compare conventional NN matching
  (NN) Eq.~(\ref{eq:zsl_nn}) versus Normalised Nearest Neighbour (NRM)
  Eq.~(\ref{eq:NRM_distance}) and Globally Corrected (GC)  matching
  Eq.~(\ref{eq:GlobalCorrectedPredict})
  (Section~\ref{sec:postprocess}). {Note that the hubness
    correction methods (NRM and GC) do not change retrieval
    performance. Therefore, NN/ NRM/ GC do not perform differently on OlympicSports and CCV.}
\item {\textbf{Transductive:} (Trans) Indicating whether the
  combination of components is transductive (\checkmark) or not
  (\text{\sffamily X}). The former requires the access to unlabelled testing data.} 
\end{itemize}

Based on this breakdown of components, we note that the condition (\text{\sffamily X}-RR-\text{\sffamily X}-NN-\text{\sffamily X}) is roughly equivalent to the methods in \citet{Socher2013} and \citet{lazaridou2014crossModalDSM}, and the conditions (\text{\sffamily X}-RR-\text{\sffamily X}-GC-\checkmark, \text{\sffamily X}-RR-\text{\sffamily X}-NRM-\checkmark) are roughly equivalent to \citet{dinu2015improving}. {We present the results in Table~\ref{tab:ComponentEval}.}

\vspace{0.1cm}\noindent\textbf{Metrics:}\quad {HMDB, UCF and USAA are classification benchmarks, so we report average accuracy metric. Olympic Sports and CCV are detection benchmarks, so we report mean average precision (mAP) metrics. We note that because distance normalization (NRM) does not change the relative rank of testing instances w.r.t. testing class, there is no difference between NRM and NN for mAP. Therefore, we insert a `$-$' for  Match-NRM on Olympic Sports and CCV. The performance for these `$-$' is the same as their NN counterparts.}

\begin{table*}[!ht]
\centering
{
\caption{Evaluation of the contribution of individual component (average \% accuracy $\pm$ standard deviation for HMDB51, UCF101 and USAA and mean average precision $\pm$ standard deviation for Olympic Sports and CCV). All `$-$' indicate no difference in performance between NN and NRM.}
\label{tab:ComponentEval}
\resizebox{0.95\textwidth}{!}{
\begin{tabular}{lccccccccc}
\toprule
\multicolumn{1}{c}{\textbf{Model}} & \textbf{Match} & \textbf{ST}         & \textbf{Data Aug}   & \textbf{Trans}      & \textbf{HMDB51}                & \textbf{UCF101} & \textbf{Olympic Sports}         & \textbf{CCV}                   & \textbf{USAA}                   \\ \hline
RR                          & NN    & X          & X          & X          & 14.5$\pm$2.7          & 11.7$\pm$1.7          & 35.7$\pm$8.8           & 20.7$\pm$3.0          & 29.5$\pm$5.5           \\
RR                          & NN    & \checkmark & X          & \checkmark & 17.0$\pm$3.1          & 15.9$\pm$2.3          & 37.3$\pm$9.1           & 21.7$\pm$3.2          & 30.2$\pm$5.2           \\
MR                          & NN    & X          & X          & \checkmark & 15.9$\pm$3.1          & 12.9$\pm$2.2          & 37.7$\pm$9.5           & 21.4$\pm$3.0          & 29.8$\pm$4.0           \\
MR                          & NN    & \checkmark & X          & \checkmark & 18.6$\pm$3.9          & 17.6$\pm$2.7          & \textbf{38.6$\pm$10.6} & \textbf{22.5$\pm$3.4} & \textbf{35.5$\pm$4.0}  \\
RR                          & GC    & X          & X          & \checkmark & 15.3$\pm$2.7          & 13.5$\pm$1.8          & 35.7$\pm$8.8           & 20.7$\pm$3.0          & 26.1$\pm$6.7           \\
RR                          & GC    & \checkmark & X          & \checkmark & 17.0$\pm$2.9          & 14.8$\pm$2.0          & 37.3$\pm$9.1           & 21.7$\pm$3.2          & 29.0$\pm$4.0           \\
RR                          & NRM   & X          & X          & \checkmark & 16.1$\pm$2.7          & 13.9$\pm$1.5          & -                      & -                     & 28.6$\pm$7.2           \\
RR                          & NRM   & \checkmark & X          & \checkmark & 17.2$\pm$2.9          & 16.1$\pm$2.2          & -                      & -                     & 28.6$\pm$7.6           \\
MR                          & NRM   & X          & X          & \checkmark & 18.0$\pm$3.2          & 15.6$\pm$2.0          & -                      & -                     & 28.2$\pm$5.4           \\
MR                          & NRM   & \checkmark & X          & \checkmark & \textbf{19.1$\pm$3.8} & \textbf{18.0$\pm$2.7} & -                      & -                     & 31.6$\pm$3.2           \\ \hline
RR                          & NN    & X          & \checkmark & X          & 20.4$\pm$2.9          & 15.7$\pm$1.6          & 38.6$\pm$7.5           & 30.3$\pm$3.9          & 28.2$\pm$4.6           \\
RR                          & NN    & \checkmark & \checkmark & \checkmark & 23.6$\pm$3.7          & 21.2$\pm$2.4          & 42.0$\pm$8.2           & \textbf{33.8$\pm$4.7} & 42.8$\pm$8.7           \\
RR                          & NRM   & X          & \checkmark & \checkmark & 21.0$\pm$2.7          & 18.5$\pm$1.7          & -                      & -                     & 35.6$\pm$2.6           \\
RR                          & NRM   & \checkmark & \checkmark & \checkmark & 23.7$\pm$3.4          & \textbf{22.2$\pm$2.6} & -                      & -                     & 42.6$\pm$9.1           \\
MR                          & NN    & X          & \checkmark & \checkmark & 20.6$\pm$2.9          & 17.2$\pm$1.6          & 41.1$\pm$7.7           & 30.4$\pm$3.9             & 30.3$\pm$4.9           \\
MR                          & NN    & \checkmark & \checkmark & \checkmark & 23.5$\pm$3.9          & 20.6$\pm$2.4          & \textbf{43.2$\pm$8.3}  & 33.0$\pm$4.8          & 41.2$\pm$9.7           \\
MR                          & NRM   & \checkmark & \checkmark & \checkmark & \textbf{24.1$\pm$3.8} & 22.1$\pm$2.5          & -                      & -                     & \textbf{43.3$\pm$10.9} \\ \bottomrule
\end{tabular}}}
\end{table*}

\vspace{0.1cm}\noindent\textbf{Experimental Results:} {We make the following observations from the results in Table~\ref{tab:ComponentEval}: (i) The simplest approach of
directly mapping features to the embedding space (\text{\sffamily
  X}-RR-\text{\sffamily X}-NN-\text{\sffamily X}
\citep{Socher2013,lazaridou2014crossModalDSM}) works reasonably well 
suggesting that semantic space is effective as a representation and
supports ZSL. (ii) Manifold regularization  reliably improves
performance compared to conventional ridge regression by reducing the
domain shift through considering the unlabelled testing data
{(transductive learning)}. (iii) Data augmentation also
significantly improves the results by providing a more representative
sample of training data for learning the embedding. (iv) In line with
previous work self-training \citep{Fu2015} and Hubness
\citep{dinu2015improving} post-processing improve results at testing
time, and this is complementary with our proposed manifold
regularization and data augmentation.
}

\subsubsection{Comparison With State-of-the-Art}
{ In addition to the above variants of our framework, we also evaluate the following state-of-the-art approaches to ZSL on action recognition tasks. As both word-vector embedding and manually labelled attributes are widely studied in the literature of zero-shot learning, we compare our approach using both word-vector and attribute semantic embedding with the state-of-the-art models. Attribute embedding is only evaluated on UCF, Olympic Sports and USAA where attributes are available.}

\vspace{0.1cm}\noindent\textbf{Word-Vector Embedding:}\quad
{
For word-vector embedding, we evaluate three alternative models:
\begin{enumerate}
\item \textbf{Structured Joint Embedding (SJE)} We use the code of \citet{akata2015outputEmbedding} with FV encoded visual feature to evaluate the performance on all 5 datasets. The SJE model employs bilinear ranking to ensure relevant labels (word-vectors) are ranked higher than irrelevant labels. 
\item \textbf{Convex Combination of Semantic Embeddings (ConSE)} We implement the ConSE model \citep{norouzi2014_ICLR} with the same FV encoded feature and evaluate on all 5 datasets. The ConSE model firstly trains classifiers for each known category $p(y_j|\vect{x})$. Given testing visual data $\vect{x}$, the semantic embedding of visual data is synthesized by a linear combination of known category embeddings as $f(\vect{x})=\sum_{j=1}^Tp(y_j|\vect{x})\vect{z}_j$ where $T$ is the top $T$ known classes.
\item \textbf{Support Vector Embedding (SVE)} Our preliminary model published in \citet{Xu_SES_ICIP15}. This model learns the visual-to-semantic mapping via support vector regression. Performance is reported on HMDB51 and UCF101 datasets.
\end{enumerate}
}

\vspace{0.1cm}\noindent\textbf{Attribute Embedding:}\quad
{
In addition to word-vector embedding based methods, we also compare against existing state-of-the-art models using attribute embeddings. To enable direct comparisons with the same embedding, we carry out experiments for our approach with attribute embedding as well (although in this setting our data augmentation cannot be applied). Specifically, we compare the following methods:
}
\begin{enumerate}
\item  \textbf{Direct Attribute Prediction (DAP)} We implement the method of \citet{Lampert2014}, but using the same FV encoded visual features and linear kernel SVM attribute classifiers $p(\vect{a}|\vect{x})$. Recognition is then performed based on attribute posteriors and manually specified attribute descriptor $p(\vect{a}|\scal{y})$.
 
\item \textbf{Indirect Attribute Prediction (IAP)} \citep{Lampert2014}. This differs from DAP by learning a per-category classifier $p(\scal{y}|\vect{x})$ from training data first and then use the training category attribute-prototype dependency $p(\vect{a}|\scal{y})$ to obtain attribute estimator $p(\vect{a}|\vect{x})$.

\item \textbf{Human Actions by Attributes (HAA)} \citep{Liu2011}. We reproduce a simplified version of this model which exploits the manually labelled attributes $\{a_m\}$ for zero-shot learning. Similar to DAP, a binary SVM classifier is trained per attribute. In the testing phase, each testing sample is projected into attribute space and then assigned to the closest testing/unknown class based on cosine distance to the class prototype (NN).

\item {\textbf{Propagated Semantic Transfer (PST)} \citep{rohrbach2013transfer} and \citep{Rohrbach2016}. Label propagation is adopted in this approach to adjust the initial predictions of DAP. Specifically, a KNN graph is constructed in the attribute embedding space and a smoothed solution is obtained transductively by semi-supervised label propagation \citep{zhou2004learning}.}

\item \textbf{Multi-Modal Latent Attribute Topic Model (M2LATM)} \citep{Fu2014a}. It exploits both user-defined and discovered latent attributes to facilitate zero-shot learning. This model fuses multiple features -- static (SIFT), motion (STIP) and audio (MFCC), and thus has an advantage compared to other methods evaluated that use vision alone.  We report the results on USAA from \citet{Fu2014a}.

{\item \textbf{Transductive Multi-View Bayesian Label Propagation (TMV-BLP) \citep{fu2014transductive}}. This model builds a common space for multiple embeddings. It combines attribute and word-vectors, and applies bayesian label propagation to infer the category of testing instances. It evaluated on USAA dataset with SIFT, STIP and MFCC features.}

{\item \textbf{Transductive Multi-View Hypergraph Label Propagation (TMV-HLP) \citep{Fu2015}}. An improved version of TMV-BLP. A distributed hypergraph was adopted to replace the local neighbourhood graph in \citet{fu2014transductive}. }

{\item \textbf{Unsupervised Domain Adaptation (UDA)}. The UDA model \citep{Kodirov2015} learns
dictionary on auxiliary data and adapts it to the target data as a constraint on
the target dictionary rather than blindly using the same dictionary.}
 \end{enumerate}
 
\vspace{0.1cm}\noindent\textbf{Mixed Embedding:}\quad
{We refer to exploiting attribute and word-vector embeddings jointly as studied by \citet{Fu2015} and \citet{akata2015outputEmbedding}. Although multi-view embedding is not the focus of this work, we evaluate our model with a simple concatenation of attribute and word-vector embeddings. Three alternatives are compared including TMV-BLP \citep{fu2014transductive}, UDA \citep{Kodirov2015} and TMV-HLP \citep{Fu2015}.}

\vspace{0.1cm}\noindent\textbf{Method Properties:}\quad
{ We indicate the nature of each approach with four parameters. \textbf{DA} - if data augmentation is applied. \textbf{Trans} - whether the approach requires transductive access to testing data. \textbf{Embed} - what semantic embedding is used. Embed-A, Embed-W and Embed A+W indicate attribute, word vector, and both attribute+word vector embeddings respectively.  \textbf{Feat} - What visual feature is used. FV indicates Fisher vector encoded dense trajectory feature; BoW indicates bag of words encoded dense trajectory feature; and SMS indicates joint SIFT, MFCC and STIP feature.
}

\begin{table*}[!htb]
\centering
\caption{Comparison with state-of-the-art approaches to ZSL. Both attribute and word-vector embeddings are studied for fair comparison. $^{*}$ performances are estimated from Fig.~2 (a) $\Gamma(X+V)$ and $\Gamma(X+A)$ respectively in \citet{fu2014transductive}. $^{**}$ performances are estimated from Fig.~5 (c) $\Gamma(X+V)$ and $\Gamma(X+A)$ respectively in \citet{Fu2015}. N/A indicates not available due to the absence of attribute annotation or not reported by the original work. }\label{tab:StateOfTheArt}
\resizebox{0.99\textwidth}{!}{
\begin{tabular}{lccccccccc}
\toprule
\multicolumn{1}{c}{\textbf{Model}}      & \textbf{DA} & \textbf{Trans} & \textbf{Embed} & \textbf{Feat}  & \textbf{HMDB51}       & \textbf{UCF101}       & \textbf{Olympic Sports} & \textbf{CCV}          & \textbf{USAA}          \\ \hline
Random Guess                            & X                 & X              & X              & X             & 4.0                   & 2.0                   & 12.5                    & 10.0                  & 25.0                   \\\hline
RR (Ours)                                     & X                 & X              & W              & FV             & 14.5$\pm$2.7          & 11.7$\pm$1.7          & 35.7$\pm$8.8            & 20.7$\pm$3.0          & 29.5$\pm$5.5           \\
MR (Ours)                                     & X                 & \checkmark     & W              & FV             & 19.1$\pm$3.8          & 18.0$\pm$2.7          & 38.6$\pm$10.6           & 22.5$\pm$3.4          & 31.6$\pm$3.2           \\
MR (Ours)                                     & \checkmark        & \checkmark     & W              & FV             & \textbf{24.1$\pm$3.8} & \textbf{22.1$\pm$2.5} & \textbf{43.2$\pm$8.3}   & \textbf{33.0$\pm$4.8} & \textbf{43.3$\pm$10.9} \\ 

SJE \citep{akata2015outputEmbedding}  & X                 & X              & W              & FV              & 12.0$\pm$2.6          & 9.3$\pm$1.7           & 34.6$\pm$7.6            & 16.3$\pm$3.1          & 21.3$\pm$0.6           \\
ConSe \citep{norouzi2014_ICLR}       & X                 & X              & W              & FV              & 15.0$\pm$2.7          & 11.6$\pm$2.1          & {36.6$\pm$9.0}   & {20.7$\pm$3.1} & {28.2$\pm$4.8}  \\
TMV-BLP \citep{fu2014transductive}$^*$                & X                 & \checkmark     & W              & SMS & N/A                   & N/A                   & N/A                     & N/A                   & {41.0}         \\
TMV-HLP \citep{Fu2015}$^{**}$                & X                 & \checkmark     & W              & SMS & N/A                   & N/A                   & N/A                     & N/A                   & {43.0}         \\
SVE \citep{Xu_SES_ICIP15}           & X        & X     & W              & BoW            & 12.9$\pm$2.3 & 11.0$\pm$1.8 & N/A                     & N/A                   & N/A                    \\ \hline
RR (Ours)                                     & X                 & X              & A              & FV             & N/A                   & 12.6$\pm$1.8          & 51.7$\pm$11.3           & N/A                   & 44.2$\pm$13.9          \\
MR (Ours)                                     & X                 & \checkmark     & A              & FV             & N/A                   & \textbf{20.2$\pm$2.2} & \textbf{53.5$\pm$11.9}  & N/A                   & \textbf{51.6$\pm$10.0} \\ 
DAP \citep{Lampert2014} & X                 & X              & A              & FV             & N/A                   & 15.2$\pm$1.9          & 44.4$\pm$9.9            & N/A                   & 37.9$\pm$5.9           \\
IAP \citep{Lampert2014} & X                 & X              & A              & FV             & N/A                   & {15.6$\pm$2.2} & 44.0$\pm$10.7           & N/A                   & 31.7$\pm$1.6           \\
HAA \citep{Liu2011}                 & X                 & X              & A              & FV             & N/A                   & 14.3$\pm$2.0          & {48.3$\pm$10.2}  & N/A                   & 41.2$\pm$9.8           \\
PST \citep{rohrbach2013transfer} 	& X                 & \checkmark              & A              & FV             & N/A                   & 15.3$\pm$2.2
          & {48.6$\pm$11.0}  & N/A                   & 47.9$\pm$10.6           \\
M2LATM \citep{Fu2014a}                & X                 & \checkmark     & A              & SMS & N/A                   & N/A                   & N/A                     & N/A                   & 41.9                   \\
TMV-BLP \citep{fu2014transductive}$^{*}$                & X                 & \checkmark     & A              & SMS & N/A                   & N/A                   & N/A                     & N/A                   & {40.0}         \\
TMV-HLP \citep{Fu2015}$^{**}$                & X                 & \checkmark     & A              & SMS & N/A                   & N/A                   & N/A                     & N/A                   & {42.0}         \\
UDA \citep{Kodirov2015}               & X                 & \checkmark     & A              & FV             & N/A                   & 13.2$\pm$1.9          & N/A                     & N/A                   & N/A                    \\ \hline
MR (Ours)                                     & X                 & \checkmark     & A+W            & FV             & N/A                   & \textbf{20.8$\pm$2.3} & \textbf{53.2$\pm$11.6}  & N/A                   & \textbf{51.9$\pm$10.1} \\
TMV-BLP \citep{fu2014transductive}                & X                 & \checkmark     & A+W            & SMS & N/A                   & N/A                   & N/A                     & N/A                   & {47.8}          \\
UDA \citep{Kodirov2015} & X                 & \checkmark     & A+W            & FV & N/A                   & 14.0$\pm$1.8                   & N/A                     & N/A                   & N/A \\
TMV-HLP \citep{Fu2015}                & X                 & \checkmark     & A+W            & SMS & N/A                   & N/A                   & N/A                     & N/A                   & {50.4}          \\ \bottomrule
\end{tabular}}
\end{table*}

\vspace{0.1cm}\noindent\textbf{Experimental Results:}\quad
The full results are presented in Table~\ref{tab:StateOfTheArt}, from
which we draw the following conclusions: (i) Our non-transductive model (RR) is strong compared with alternative models with either word-vector embedding or attribute embedding. For example, our RR model is able to beat SJE and ConSE in UCF101, CCV and USAA with word-vector embedding and beat DAP, IAP and HAA in Olympic Sports and USAA. (ii) With transductive access to testing data, our model MR-X-\checkmark-W is better than most alternative models with word-vector and competitive against models with attribute embedding. (iii) The overall combination of all components, manifold regularized regression (MR), Data Augmentation (DA) and Self-training and hubness (Trans), with word-vector embedding (MR-\checkmark-\checkmark-W) can yield very competitive performance. Depending on the
dataset, our overall model is comparable or significantly better than the
attribute-centric methods, e.g. UCF101. (iv) With mix-embedding (A+W) our model is still very competitive against existing ZSL approaches and outperform TMV-BLP, UDA and TMV-HLP. Apart from the above observations we note that the ZSL
  performance variance is relatively high, particularly in Olympic
  Sports and USAA datasets. This is because specific choice of
  train/test classes in ZSL matters more than specific choice of
  train/test instances in conventional supervised learning. E.g., in
  olympic sports there are highly related categories `high jump' - `long jump' and  `diving platform 10m' - `diving springboard 3m'. Recognition performance is higher when these pairs are separated in training and testing, and lower if they are both in testing. This issue is explored further in Sec.~\ref{sec:dataAug}

\subsubsection{Generalising the Transductive Setting}
{
In this section, we study the possibility to apply the transductive learning ideas investigated here to improve existing zero-shot learning approaches with both word-vector and attribute embeddings. In particular we consider transductive generalisations of three alternative models SJE, ConSE and HAA.
}

\vspace{0.1cm}\noindent\textbf{SJE: }\quad 
{
SJE \citep{akata2015outputEmbedding} uses a bi-linear mapping to evaluate the compatibility between novel instances and word-vectors. Suppose we have the bi-linear form $\vect{x}^\top\matr{W}\vect{z}$ to compute the compatibility score between category name word-vector $\vect{z}$ (output embedding) and video instance $\vect{x}$ (input embedding) which corresponds to Eq~(1) in \citet{akata2015outputEmbedding}. Given learned model $\matr{W}$ we can first project video instance by this mapping as $\vect{x}^\top\matr{W}$. Then we can apply self-training to adjust the novel category's output embedding $\vect{z}$ as,
\begin{equation}
\tilde{\vect{z}}=\frac{1}{|NN_k(\vect{z})|}\sum\limits_{i\in NN_k(\vect{z})}(\vect{x}_i^\top\matr{W})
\end{equation}
where the function $NN_k(\cdot)$ returns the k nearest neighbour of $\vect{z}$ w.r.t. all testing video instances $\{\vect{x}_i^\top \matr{W}\}$. The adjusted category embedding replaces the original output embedding for prediction. We can resolve the hubness issue for bi-linear model as well. Specifically, we use the $1-\vect{x}^\top\matr{W}\vect{z}$ normalised to between 0 and 1 as the distance and apply the same distance normalization trick introduced in Eq~(\ref{eq:NRM_distance}).
}

\vspace{0.1cm}\noindent\textbf{ConSE:}\quad
{
We train SVM classifiers for each known category as $p(y_j|\vect{x})$ and take the top $T$ responses for a testing instance to synthesize the embedding as,
 \begin{equation}
 f(\vect{x_i})=\frac{1}{T}\sum\limits_{j=1}^T p(y_j|\vect{x}_i) \vect{z}_j
 \end{equation} 
 where $\matr{z}_j$ is the semantic embedding of $j$-th known category. To apply self-training, we simply do the same calculation w.r.t. embeddings of testing videos as,
 \begin{equation}
 \tilde{\vect{z}}=\frac{1}{|NN_k(\vect{z})|}\sum\limits_{i\in NN_k(\vect{z})}f(\vect{x}_i)
 \end{equation}
 Hubness correction can be integrated in the same way.
}

\vspace{0.1cm}\noindent\textbf{HAA:}\quad
{
We do nearest neighbour matching in attribute embedding space, so both self-training and hubness correction can be applied in the same ways as our model.  }
 
 \begin{table*}[]
\centering
{\caption{Study the possibility to generalize transductive settings to existing zero-shot learning approaches.}
\label{tab:TransStudy}
\begin{tabular}{lcccccccccc}
\toprule
\multicolumn{1}{c}{Model} & Match & ST         & Trans      & Embed & Feat & HMDB51       & UCF101       & Olympic Sports & CCV          & USAA          \\ \hline
SJE                       & NN    & X          & X          & W     & FV   & 12.0$\pm$2.6 & 9.3$\pm$1.7  & 34.6$\pm$7.6   & 16.3$\pm$3.1 & 21.3$\pm$0.6  \\
SJE                       & NN    & \checkmark & \checkmark & W     & FV   & 10.5$\pm$2.4 & 8.9$\pm$2.2  & 32.5$\pm$6.7   & 15.4$\pm$3.1 & 27.7$\pm$7.1  \\
SJE                       & NRM   & X          & \checkmark & W     & FV   & 12.7$\pm$2.4 & 10.5$\pm$1.7 & -              & -            & 19.8$\pm$6.7  \\
SJE                       & NRM   & \checkmark & \checkmark & W     & FV   & 10.6$\pm$2.3 & 9.2$\pm$2.0  & -              & -            & 26.8$\pm$9.2  \\ \hline
ConSE                     & NN    & X          & X          & W     & FV   & 15.0$\pm$2.7 & 11.6$\pm$2.1 & 36.6$\pm$9.0   & 20.7$\pm$3.1 & 28.2$\pm$4.8  \\
ConSE                     & NN    & \checkmark & \checkmark & W     & FV   & 15.4$\pm$2.8 & 12.7$\pm$2.2 & 37.0$\pm$9.9   & 21.2$\pm$3.1 & 28.3$\pm$4.2  \\
ConSE                     & NRM   & X          & \checkmark & W     & FV   & 15.8$\pm$2.6 & 12.7$\pm$2.1 & -              & -            & 26.2$\pm$9.5  \\
ConSE                     & NRM   & \checkmark & \checkmark & W     & FV   & 16.3$\pm$3.1 & 12.9$\pm$2.2 & -              & -            & 26.3$\pm$9.4  \\ \hline
HAA                       & NN    & X          & X          & A     & FV   & N/A          & 14.3$\pm$2.0 & 48.3$\pm$10.2  & N/A          & 41.2$\pm$9.8  \\
HAA                       & NN    & \checkmark & \checkmark & A     & FV   & N/A          & 18.7$\pm$2.4
 & 49.4$\pm$10.8  & N/A          & 47.6$\pm$10.5 \\
HAA                       & NRM   & X          & \checkmark & A     & FV   & N/A          & 15.9$\pm$1.9 & -              & N/A          & 48.4$\pm$8.9  \\
HAA                       & NRM   & \checkmark & \checkmark & A     & FV   & N/A          & 19.1$\pm$2.3
 & -              & N/A          & 49.4$\pm$9.0  \\ \bottomrule
\end{tabular}}
\end{table*}

\vspace{0.1cm}\noindent\textbf{Experimental Results:} {The results on generalizing other methods to the transductive setting are presented in Table~\ref{tab:TransStudy}. We observe that hubness correction (NRM)  improves performance on HMDB51 and UCF101 for all three models. The effect is not so clear on USAA except for HAA. As hubness correction does not change the rank of individual testing instances w.r.t. testing category, no improvement is expected on Olympic Sports and CCV from NN to NRM. Self-training is in general effective for ConSE and HAA but is detrimental to SJE. This may be due to SJE's ranking loss: It aims to rank, rather than project video instances to the vicinity of their category embedding. Therefore, the projected video instances ($\vect{x}^\top\matr{W}$) do not form neat clusters in the word-vector space which makes self-training ineffective.}

\subsection{Zero-shot Learning of Complex Events}

In this section, we experiment on the more challenging complex event dataset - TRECVID MED 2013.

\vspace{0.1cm}\noindent\textbf{Data Split:}\quad 
\textcolor{black}{We study the 30 classes of the MED test set, holding out the 20 events specified by the 2013 evaluation scheme for zero-shot recognition, and training on the other 10. We train on the total 1611 videos in Event Kit Train (160 per event in average) and test on the 27K examples in MED test, of which only about 1448 videos are the 20 events to be detected. This is different to the standard TRECVID MED 2013 0EK in which concept detectors are trained on the Research Set \citep{Habibian2014,habibian2014composite,Wu2014}}.
{This experimental design is chosen because we want to exploit
  {\em only} per-category annotation (event name) as semantic
  supervision, rather than requiring the per-video sentence annotation
  used in the Research Set which is very expensive to collect. {We note that with few exceptions \citep{Jain} TRECVID MED 2013 is rarely addressed with event name annotation only.} With this assumption, it means we use fewer training videos (1611)  compared to the 10K video Research Set. Thus our results are not
  comparable to existing TRECVID MED 2013 0EK benchmark results,
  because we use  vastly less training data.} 

\begin{table}[h]
\centering
\caption{Events for training visual to semantic regression}
\resizebox{0.5\textwidth}{!}{
\begin{tabular}{l|l|l|l}
\hline
{\bf ID} & {\bf Event Name}                 & {\bf ID} & {\bf Event Name}           \\ \hline
E001     & Attempting a board trick         & E002     & Feeding an animal          \\ \hline
E003     & Landing a fish                   & E004     & Wedding ceremony           \\ \hline
E005     & Working on a woodworking project & E016     & Doing homework or studying \\ \hline
E017     & Hide and seek                    & E018     & Hiking                     \\ \hline
E019     & Installing flooring              & E020     & Writing                    \\ \hline
\end{tabular}\label{tab:TRECVID_train_event}}
\end{table}

\begin{table}[h]
\centering
\caption{Events for testing zero-shot event detection}
\resizebox{0.5\textwidth}{!}{
\begin{tabular}{l|l|l|l}
\hline
{\bf ID} & {\bf Event Name}                 & {\bf ID} & {\bf Event Name}                  \\ \hline
E006     & Birthday party                   & E007     & Changing a vehicle tire           \\ \hline
E008     & Flash mob gathering              & E009     & Getting a vehicle unstuck         \\ \hline
E010     & Grooming an animal               & E011     & Making a sandwich                 \\ \hline
E012     & Parade                           & E013     & Parkour                           \\ \hline
E014     & Repairing an appliance           & E015     & Working on a sewing project       \\ \hline
E021     & Attempting a bike trick          & E022     & Cleaning an appliance             \\ \hline
E023     & Dog show                         & E024     & Giving directions to a location   \\ \hline
E025     & Marriage proposal                & E026     & Renovating a home                 \\ \hline
E027     & Rock climbing                    & E028     & Town hall meeting                 \\ \hline
E029     & Winning a race without a vehicle & E030     & Working on a metal crafts project \\ \hline
\end{tabular}\label{tab:TRECVID_test_event}}
\end{table}

\vspace{0.1cm}\noindent\textbf{Baselines:}\quad We compare 5 alternative baselines for TRECVID MED zero-shot event detection. 
\begin{enumerate}
\item \textbf{Random Guess} - Randomly rank the candidates.
\item \textbf{NN} (\text{\sffamily X}-RR-\text{\sffamily X}-NN-\text{\sffamily X}). Rank videos with $l_2$ distance in the semantic space.
\item \textbf{NN + ST} (\text{\sffamily X}-RR-\checkmark-NN-\checkmark). Adjust prototypes with self-training.
\item \textbf{Manifold} (\text{\sffamily X}-MR-\text{\sffamily X}-NN-\checkmark). Add manifold regularization term in the visual to semantic regression model.
\item \textbf{Manifold + ST} (\text{\sffamily X}-MR-\checkmark-NN-\checkmark) - manifold regularization regression with self-training.
\end{enumerate}



\noindent {We were not able to investigate data augmentation for TRECVID due to the different feature encoding from the other action datasets.}

\begin{figure*}[!htb]
\centering
\includegraphics[width=0.82\linewidth]{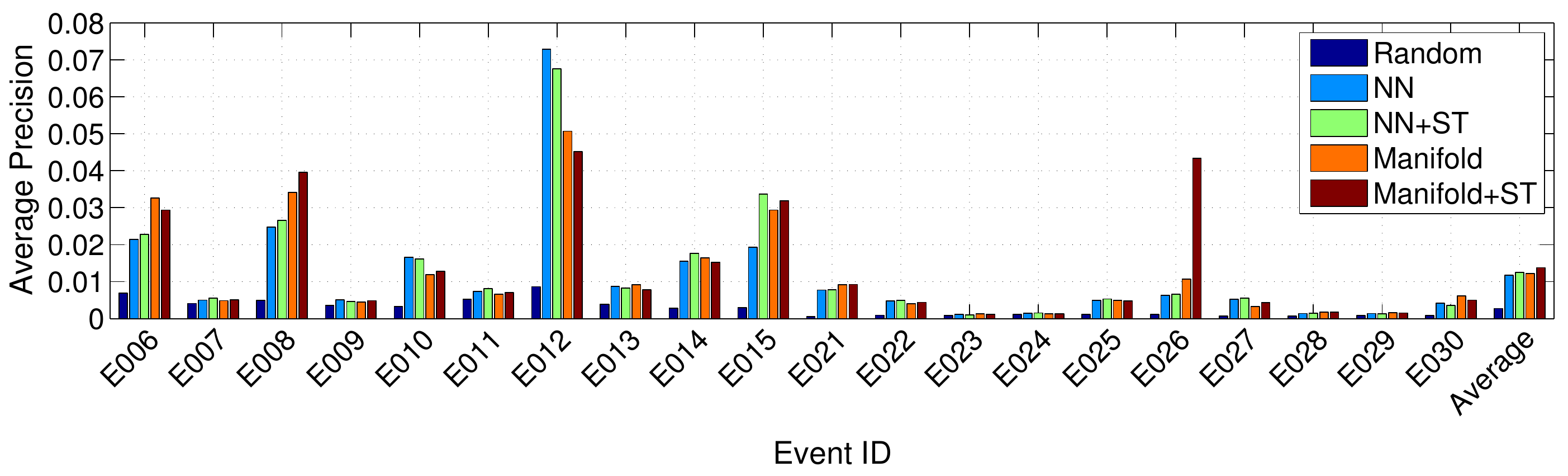}
\caption{Zero-shot performance on TRECVID MED 2013 measured in mean average precision (mAP).
}\label{fig:AP_TRECVID}
\end{figure*}

We present the performance of zero-shot learning on TRECVID
  MED 2013 in Fig.~\ref{fig:AP_TRECVID} and
  Table~\ref{tab:ComparisonTrecvid}. Fig.~\ref{fig:AP_TRECVID} reports
  the performance of 4 alternative models and random guess baseline in detecting 20 events in
  mean average precision (mAP) and the average over all events
  (Average). Compared to Random Guess ($0.28\%$), our direct embedding
  approach (NN) is  effective at zero-shot video
  detection. Self-Training and Manifold Regularization further improve
  the performance. Table~\ref{tab:ComparisonTrecvid} puts the results
  in broader context by summarising them in terms of absolute
  performance.

\begin{table}[]
\centering
{\caption{Event detection performance on TRECVID MED 2013. mAP across 20 events to be detected.}
\label{tab:ComparisonTrecvid}
\begin{tabular}{c|c|c|c}
\toprule
\textbf{Embed} & \textbf{ST}          & \textbf{Match} & \textbf{Average mAP} \\ \hline
RR             & \text{\sffamily X} & NN             &      $1.18\%$      \\
RR             & \checkmark           & NN             &    $1.25\%$         \\
MR             & \text{\sffamily X} & NN             &       $1.22\%$      \\
MR             & \checkmark           & NN             &       $1.38\%$        \\ \hline
\multicolumn{3}{c|}{Random Guess} & $0.28\%$\\
\bottomrule
\end{tabular}}
\end{table}

\subsection{Zero-Shot Qualitative Visualization}\label{sec:ZSL_Qualitative}
\begin{figure*}[!ht]
\centering
\vspace{-0.2cm}
\subfigure[Category set 1]{\includegraphics[width=.85\linewidth]{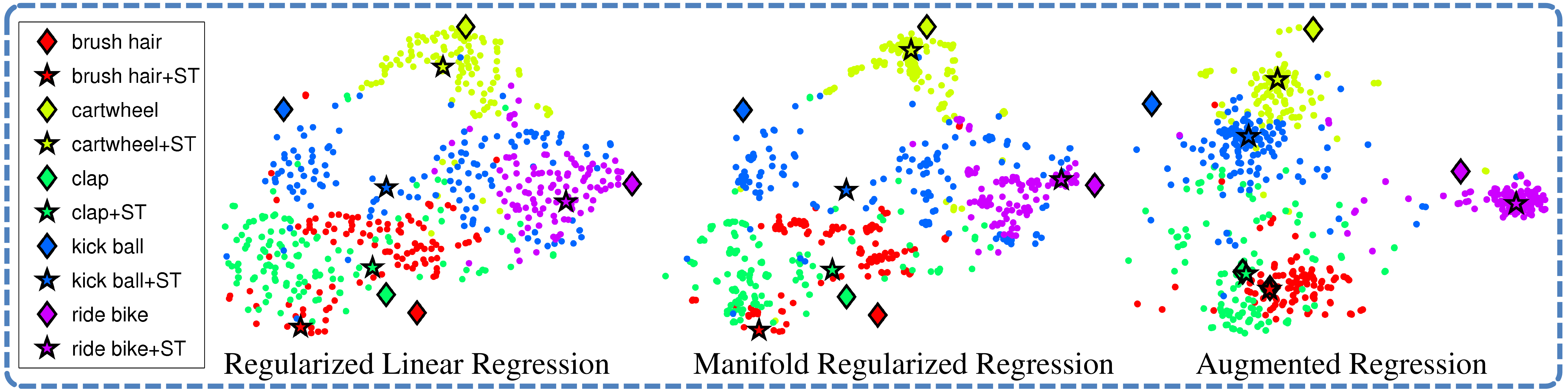}}\\
\vspace{-0.5cm}
\subfigure[Category set 2]{\includegraphics[width=.85\linewidth]{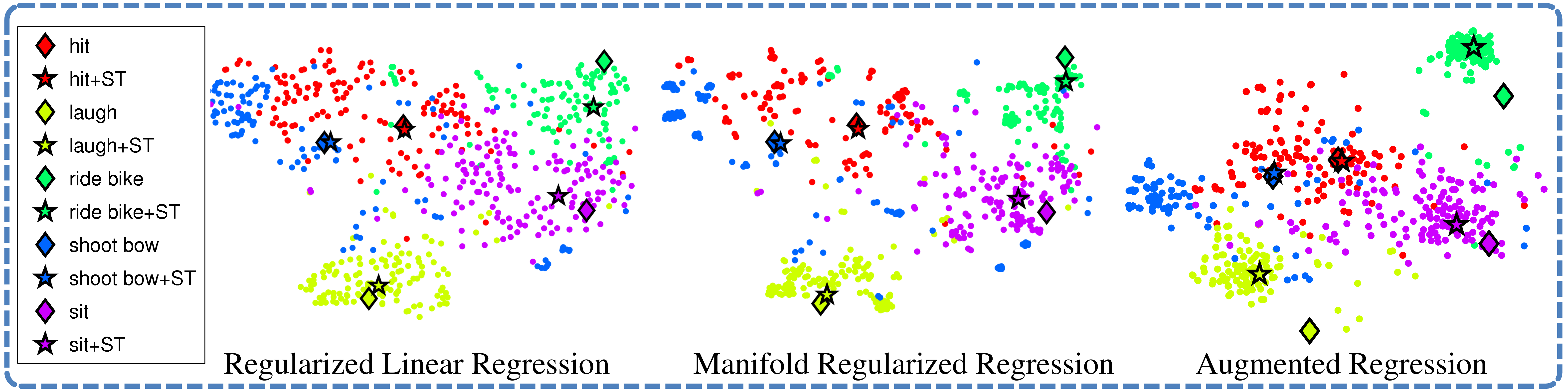}}\\
\vspace{-0.5cm}
\subfigure[Category set 3]{\includegraphics[width=.85\linewidth]{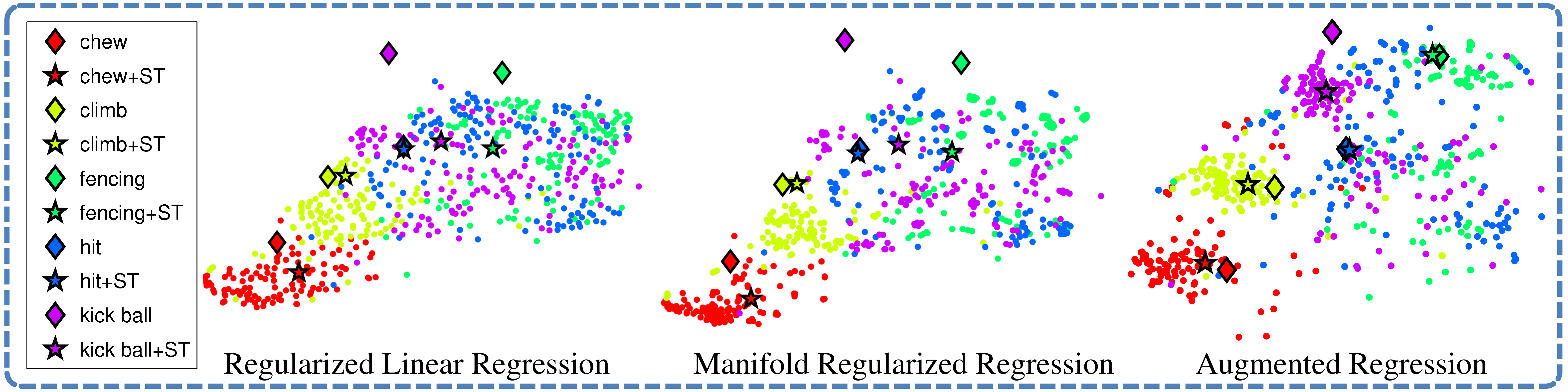}}
\vspace{-0.5cm}
\caption{A qualitative t-SNE illustration of ZSL with semantic space representation for random testing class subsets (a), (b) and (c). Variants: ridge regression, manifold regression and data augmented manifold regression. Dots indicate instances, color categories, and star/diamond show category prototypes  with/without self-training.}
\label{fig:TSNE_visualization}
\end{figure*}

\textcolor{black}{In this section we illustrate qualitatively the effect of our contributions on the resulting embedding space matching problem. For visualisation, we randomly sample 5 testing classes from HMDB51 and project all samples from these classes into the semantic space by (i) conventional ridge regression; (ii) manifold regularized regression and (iii) manifold  regularized ridge regression with data augmentation. The results are visualised in 2D in Fig.~\ref{fig:TSNE_visualization} with t-SNE \citep{Maaten2008}. Three sets of testing classes are presented for diversity.  Data instances are shown as dots, prototypes (class name projections) as diamonds, and self-training adapted prototypes as stars. Colours indicate category. }

There are three main observations from Fig.~\ref{fig:TSNE_visualization}: (i) Manifold regularized regression  yields better visual semantic projections as instances of the same class tend to form tighter clusters. This is due to the constraint of preserving the manifold structure from the visual feature space. (ii) Data augmentation yields an even more accurate projection of unseen data, as instances are projected closer to the prototypes and classes are more separable. (iii) Self-training is effective as the adapted prototypes (stars) are  closer to the center of the corresponding samples (dots) than the original prototypes (diamonds). These observations illustrate the mechanism of our ZSL accuracy improvement on conventional approaches.

{These qualitative illustrations also give intuition about why the previous result in Fig~\ref{fig:AP_TRECVID} is one of a moderate overall increase in mean AP that is the result of a varied impact of the AP for individual classes. Depending on the data and initial prototype positions, the self-training sometimes makes a very effective adjustment to the prototypes, and other times it makes little adjustment to the prototype, and hence that class' AP. E.g., In Fig.~\ref{fig:TSNE_visualization}(a), Augmented: compare blue/yellow classes versus red class. }

\subsection{Understanding ZSL and Predicting Transferrability}\label{sec:dataAug}

In this section we present further insight into considerations on what
factors will affect the efficacy of ZSL, through a category-level analysis.
The basic assumption of ZSL is that the embedding $f(\vect{x})$ trained on
known class data, will also apply to testing classes. As we have
discussed throughout this study, this assumption is stretched to some extent due to the
disjoint training and testing category sets. This leads us to investigate
how zero-shot performance depends on the specific choice of training
classes and their relation to the held out testing classes. 

\vspace{0.1cm}\noindent\textbf{Impact of training class choice on
  testing performance: } We first investigate whether there are
specific classes which, if included as training data, significantly
impact testing class performance. To study this, we compute the
correlation between training class inclusion and testing
performance. {Specifically, we consider a pair of random
  variables $\{b^{tr}_i,e^{te}_j\}$ where $b^{tr}_i$ is a binary
  scalar indicating if the $i$th class is in the training set and
  $e_j$ is the recognition accuracy of the $j$th testing
  class}. We compute the correlation $corr(i,j)$ between every pair of
variables over the 50 random splits: 

\begin{equation}\label{eq:Correlation}
corr(i,j) = \frac{\mathbb{E}[({b}^{tr}_i-\overline{{b}^{tr}_i})({e}^{te}_j-\overline{e^{te}_j})]}{var(b^{tr}_i)var(e^{te}_j)}.
\end{equation}

\begin{figure*}[!htb]
\centering
\subfigure[HMDB51 class correlation]{\includegraphics[width=0.39\linewidth]{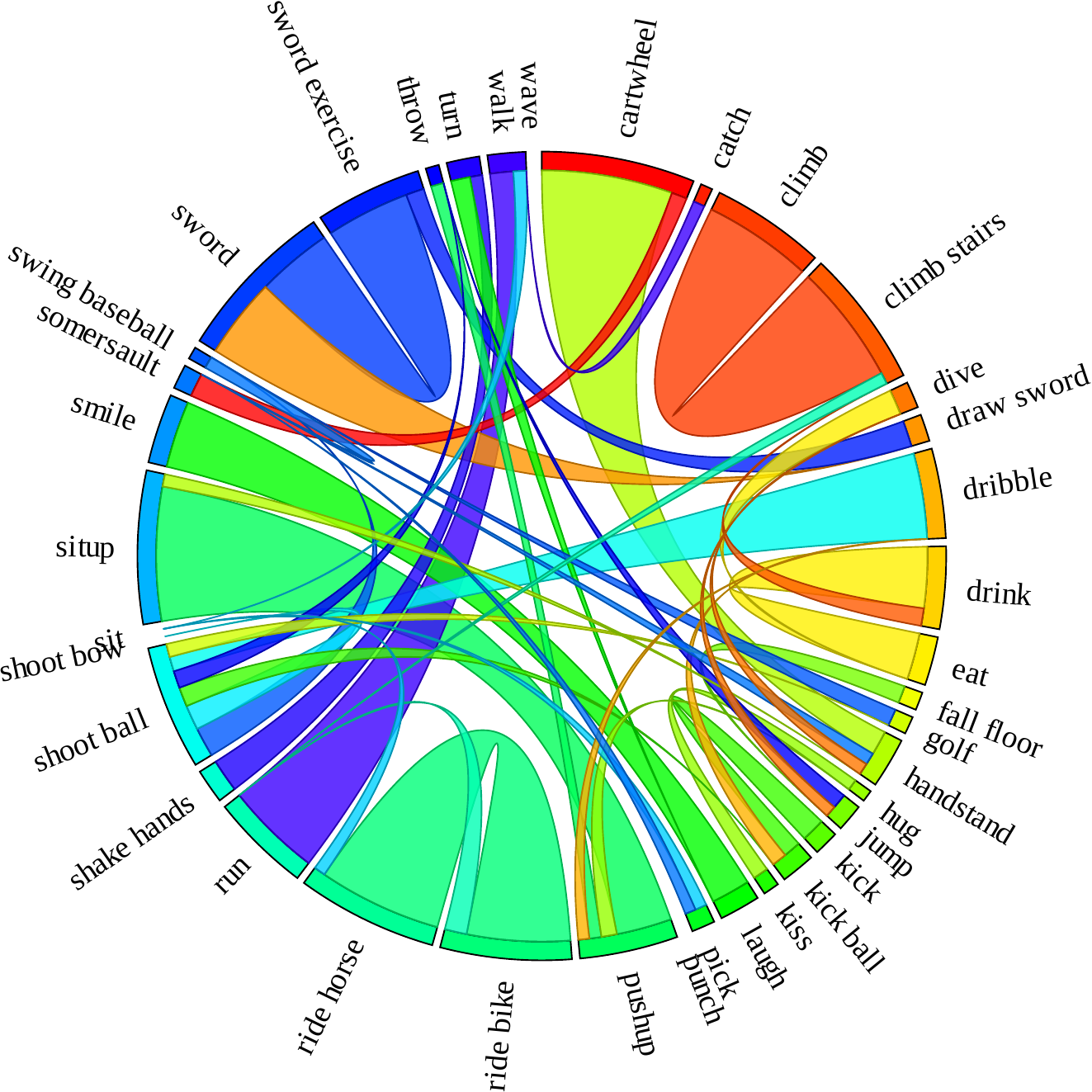}} \vspace{-0.5cm}
\subfigure[HMDB51 class name affinity]{\includegraphics[width=0.39\linewidth]{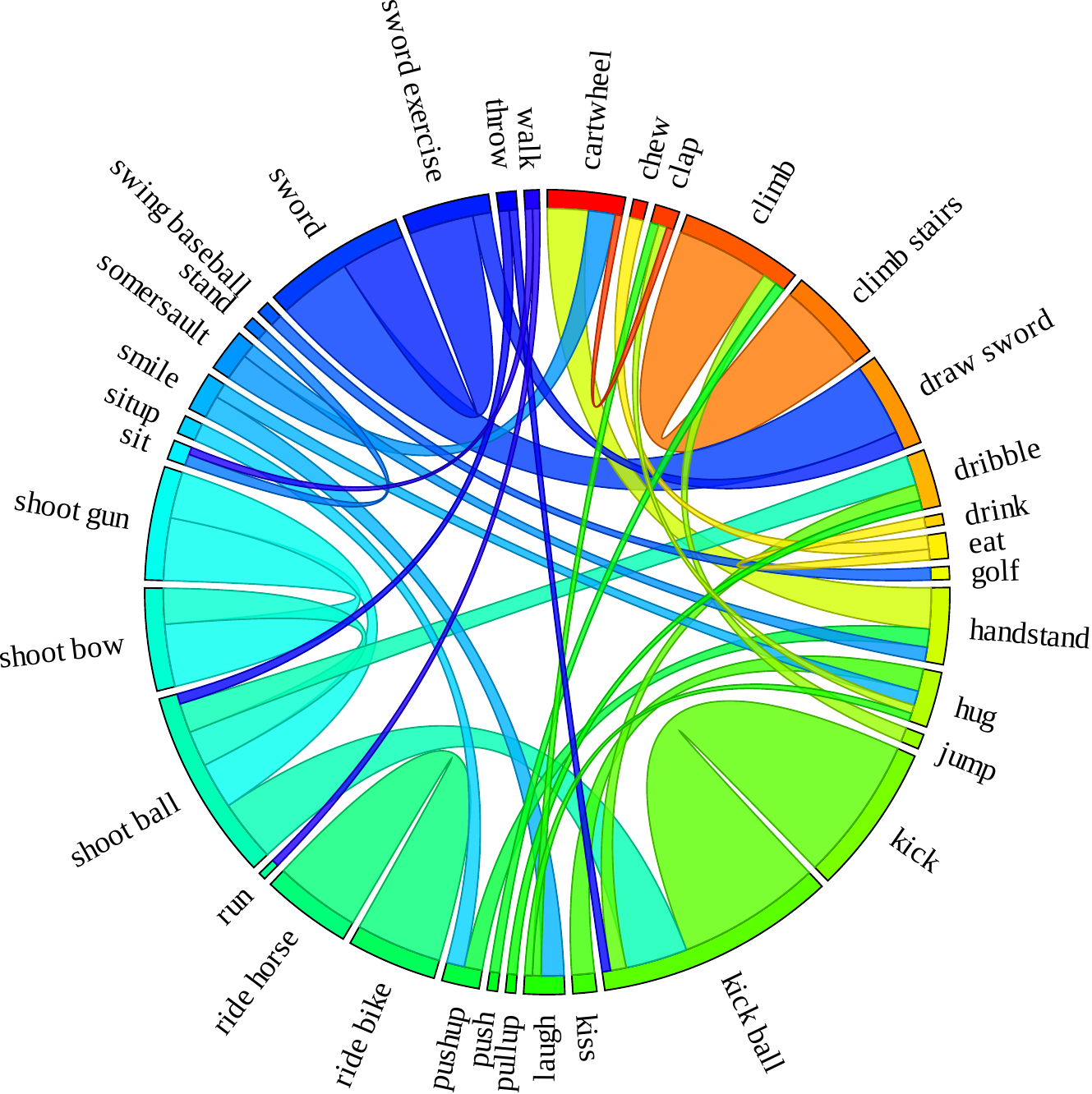}}
\subfigure[Olympic Sports class correlation]{\includegraphics[width=0.39\linewidth]{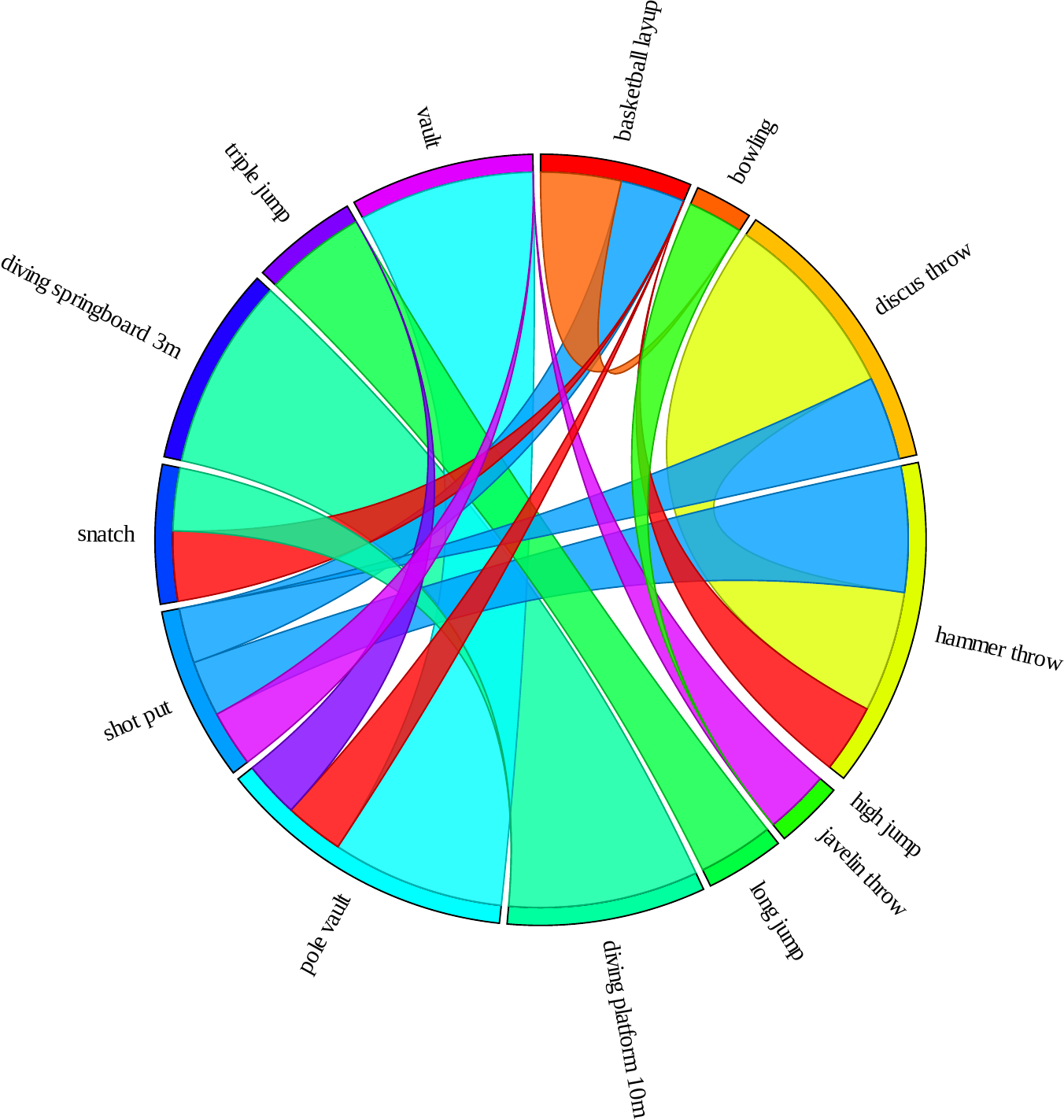}}
\subfigure[Olympic Sports class name affinity]{\includegraphics[width=0.39\linewidth]{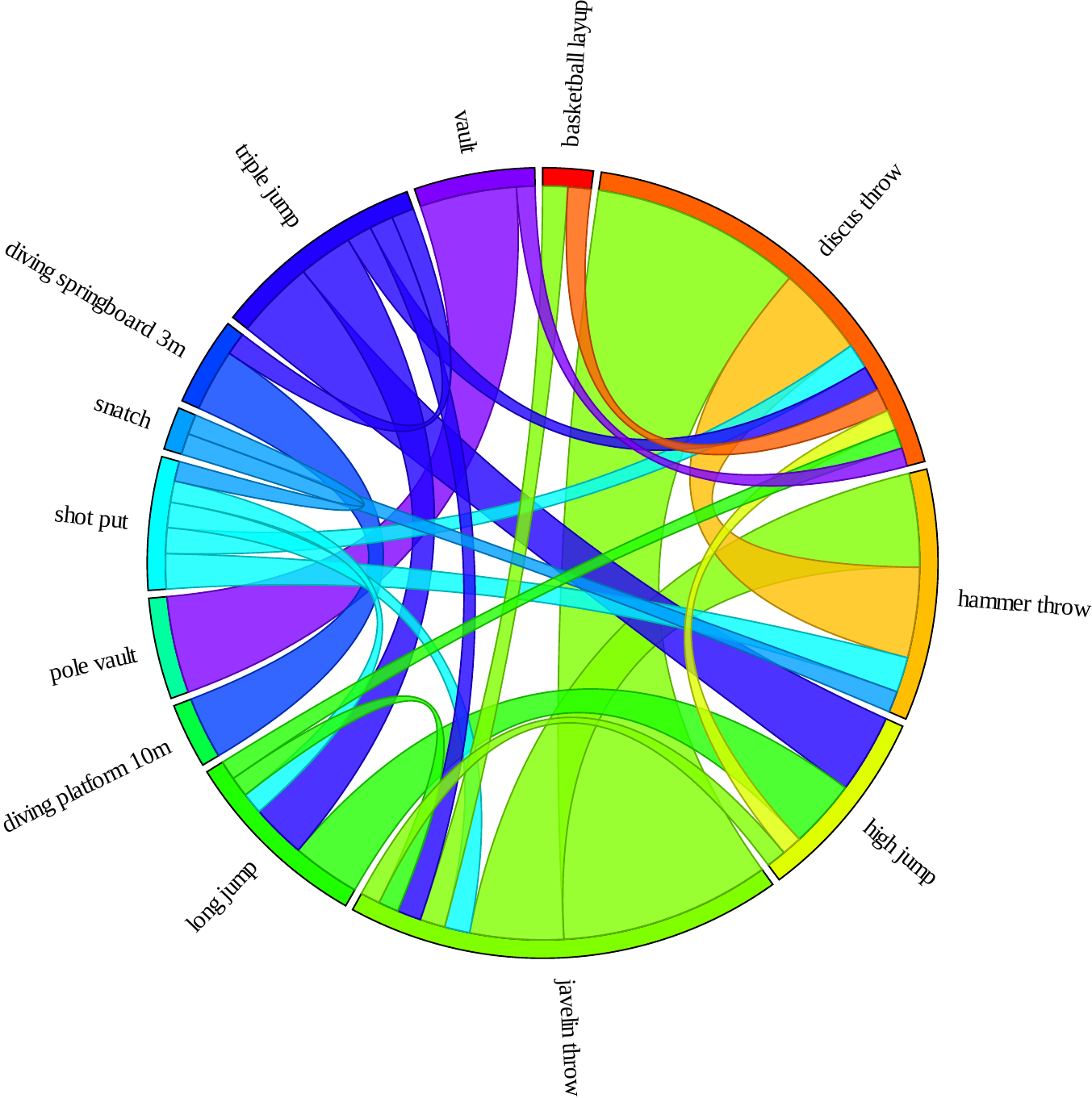}}
\caption{Chord Diagram to illustrate the category correlation discovered from zero-shot recognition experiments. (a) and (c) illustrate the correlation discovered from 50 random split zero-shot experiments; (b) and (d) illustrate the class name affinity in word-vector embedding space measured as cosine similarity.}\label{fig:CorrChordDiagram}
\end{figure*}

We use chord diagrams to visualize the relation between categories in Fig~\ref{fig:CorrChordDiagram}(a). The strength of positive cross-category correlation is indicated by the width of the bands connecting the categories on the circle. I.e., a wide band indicates inclusion of one category as training data facilitates the zero-shot recognition of the other\footnote{Due to the large number of categories we apply two preprocessing steps before plotting: (1) Convert all correlation coefficients to positive value by exponentially power scaling the correlation coefficient; (2) Remove highly negative correlated pairs to avoid clutter.}.

{We can make several qualitative observations from the chord diagrams. The class correlation captures the dependence of category B's recognition rate on category A's presence in the training set. So for instance for A=\textit{ride horse} and B=\textit{ride bike},  Fig.~\ref{fig:CorrChordDiagram}(a) shows that  we would expect high recognition accuracy of \textit{ride horse} if \textit{ride bike} is present in training set and vice versa. However while \textit{cartwheel}  supports the recognition of \textit{handstand}, the reverse is not true.}

\vspace{0.1cm}\noindent\textbf{Cross-class transferability correlates with word-vector similarity: } 
{We next investigate the affinity between class names' vector
  representations, and cross-class transferability. Class name
  affinities are shown in Fig~\ref{fig:CorrChordDiagram}(b) as chord
  diagrams. Visually there is some similarity to the cross-class
  transferability presented in Fig~\ref{fig:CorrChordDiagram}(a). To
  quantify this connection between transfer efficacy and classname
  relatedness, we vectorise the correlation
  (Fig~\ref{fig:CorrChordDiagram}(a)) and class name affinity
  (Fig~\ref{fig:CorrChordDiagram}(b)) matrices ($51\times 51$) into
  2601 dim vectors and then compute the correlation coefficients
  between the two vectors. The correlation is $0.548$, suggesting that
  class name relatedness and efficacy for ZSL are indeed
  connected. This is to say, if class A is present in training set and
  class B in testing set, and A has high affinity with B in
  word-vector distance measure, we could expect high performance in
  recognizing class B.} 

To qualitatively illustrate this connection, we list the top 10 positively correlated category pairs in Table~\ref{tab:PosCorrPairs}. Here the correlation of action 1 being in training and action 2 in testing is given as \textit{Fwd Corr}, with \textit{Back Corr} being the opposite. The affinity between category names are given as \textit{WV Aff} which is defined as percentile rank of word-vector distance (closer to 1 means more similar). Clearly highly correlated categories have higher word-vector similarity. 
\begin{table}[!htb]
\centering
\caption{Top 10 positive correlated class pairs}
\label{tab:PosCorrPairs}
\resizebox{0.49\textwidth}{!}{
\begin{tabular}{l|l|l|l|l}
\toprule
Action 1       & Action 2   & Fwd Corr & Back Corr & WV Aff \\ \hline
climb stairs   & climb      & 0.94     & 0.92      & 0.98        \\ \hline
ride horse     & ride bike      & 0.95     & 0.91      & 0.98        \\ \hline
situp          & pushup     & 0.96     & 0.79      & 0.91        \\ \hline
sword exercise & sword      & 0.87     & 0.85      & 0.98        \\ \hline
handstand      & cartwheel  & 0.62     & 0.96      & 0.97         \\ \hline
eat            & drink      & 0.75     & 0.81      & 0.96        \\ \hline
smile          & laugh      & 0.82     & 0.72      & 0.97       \\ \hline
walk           & run        & 0.61     & 0.90      & 0.96         \\ \hline
shoot ball     & dribble    & 0.52     & 0.87      & 0.97        \\ \hline
sword          & draw sword & 0.86     & 0.45      & 0.98        \\ \bottomrule
\end{tabular}}
\end{table}

\begin{figure}[!h]
\centering
\includegraphics[width=0.8\linewidth]{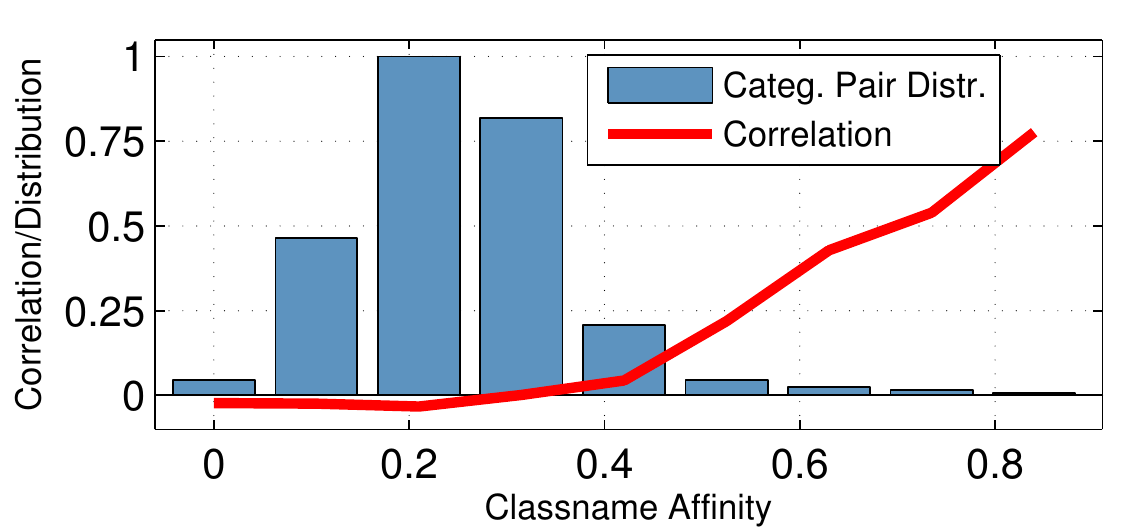}
\caption{The connection between transfer efficacy and classname affinity: Illustrated by class correlation v.s. class name affinity.}\label{fig:CorrVsAff}
\end{figure}

Although zero-shot transfer overall is effective, there are also some individual negative correlations. We illustrate the distribution of positive and negative transfer outcomes in Fig.~\ref{fig:CorrVsAff}. Here we sort all the class pairings into ten bins by their name affinity and plot the resulting histogram (blue bars). Clearly the majority of pairs have low classname affinity. For each bin of class-pairs, we also compute their average correlation defined in Eq~\ref{eq:Correlation} (Fig.~\ref{fig:CorrVsAff}, red line). There are a few observations to be made: (i) Class name affinity is clearly related to positive correlation: the correlation (red line) goes up significantly for high-affinity class pairs. (ii) There are a relatively small number of category pairs that account for the high positive correlation outcomes (low blue bars to the right). This suggests that overall ZSL efficacy is strongly impacted by the presence of key supporting classes in the training set. (iii) There are a larger number of category pairs which exhibit negative transferability (red correlation is negative around affinity of 0.2). However negative transfer effects are quantitatively  weak compared to positive transfer  (red correlation line gets only weakly negative but strongly positive).

\vspace{0.1cm}\noindent\textbf{Predicting Transferability: } 
Based on the previous observations we hypothesize that class name
affinity is predictive of ZSL performance, and may provide a guide to
selecting a good set of training classes to maximise  ZSL
efficacy. This is desirable in real application as it is often beneficial to best utilize the limited availability to annotate most useful training data for the recognition of novel categories. {We formally define the problem as given fixed
  testing categories $\{\scal{y}_j|\scal{y}_j\in \vect{y}_{te}\}$, we find the $S\%$ subset
  of  training categories $\{\scal{y}_i|\scal{y}_i\in \vect{y}_{tr}\}$ which maximize the
  performance of recognizing testing classes based on their affinity
  to the testing classes. We first of all explore three alternative
  (point-to-set) distances to measure the affinity of each training
  class $\scal{y}_i$ to the set of testing classes $\{\scal{y}_j| \scal{y}_j\in \vect{y}_{te}\}$,
  specifically the maximal/mean/minimal class name affinity:} 

{
\begin{equation}\label{eq:ClassRelatedness}
\begin{split}
&R_{max}(\scal{y}_i,\vect{y}_{te})=\underset{\scal{y}_j \in \vect{y}_{te}}{\mathrm{max}} \left(1-||g(\scal{y}_i)-g(\scal{y}_j)||\right)\\
&R_{mean}(\scal{y}_i,\vect{y}_{te})=\underset{\scal{y}_j \in \vect{y}_{te}}{\mathrm{mean}} \left(1-||g(\scal{y}_i)-g(\scal{y}_j)||\right)\\
&R_{min}(\scal{y}_i,\vect{y}_{te})=\underset{\scal{y}_j \in \vect{y}_{te}}{\mathrm{min}} \left(1-||g(\scal{y}_i)-g(\scal{y}_j)||\right)
\end{split}
\end{equation}
}

\noindent These metrics provide a plausible means to quantify the relevance of any potential training class to the testing set.  We  explore their ability to predict transferability and hence construct a good training set for a particular set of testing classes.

For this experiment, we use HMDB51 with the same 50 random  splits
introduced in Section~\ref{sect:exp_ZSL}. Keeping the testing sets
fixed, we train two alternative models based on different subsets of
each training split. Specifically: (1) \textit{Related Model} selects
the top $S\%$ most related training classes (high affinity measure by
$R(\scal{y}_i,\vect{y}_{te})$) to the testing set defined by relatedness measure in
Eq.~(\ref{eq:ClassRelatedness}) in order to learn the mapping; while
(2) \textit{Unrelated Model} selects the most $100-S\%$
unrelated. Fig.~\ref{fig:RelatedModelPerformance} shows the
performance of both models as $S$ varies between 0 and 100, where
\textit{Related} selects the top $S\%$ and \textit{Unrelated} the
bottom $100-S\%$. {Note that when $S=0\%$ and $S=100\%$ the
  \textit{Unrelated} and \textit{Related} models both select all
  training classes. Both are then equivalent to the standard ZSL model
  \text{\sffamily X}-RR-\text{\sffamily X}-NN-\text{\sffamily X}
  introduced in Table~\ref{tab:ComponentEval}. We illustrate the
  performance of both models and three alternative training-to-testing
  affinity measures in Fig.~(\ref{fig:RelatedModelPerformance}).} 

{
\begin{figure}[!htb]
\centering
\subfigure[Maximal class name affinity]{\includegraphics[width=0.8\linewidth]{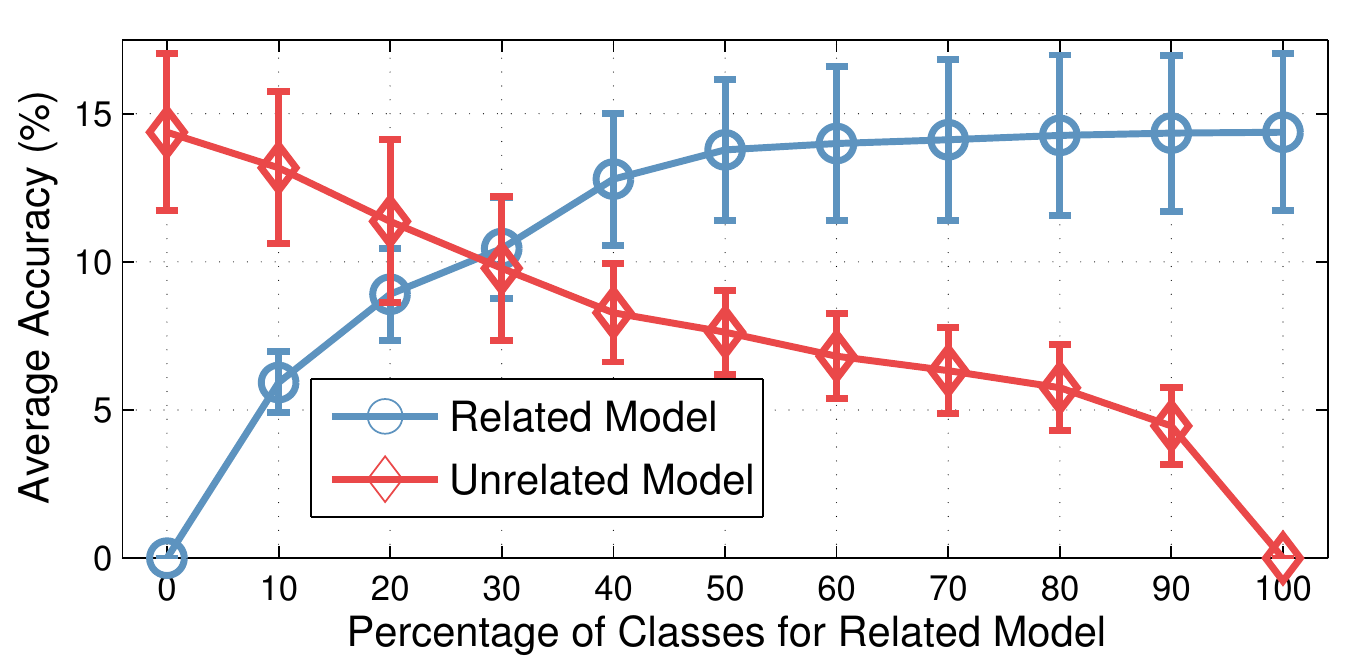}}
\subfigure[Mean class name affinity]{\includegraphics[width=0.8\linewidth]{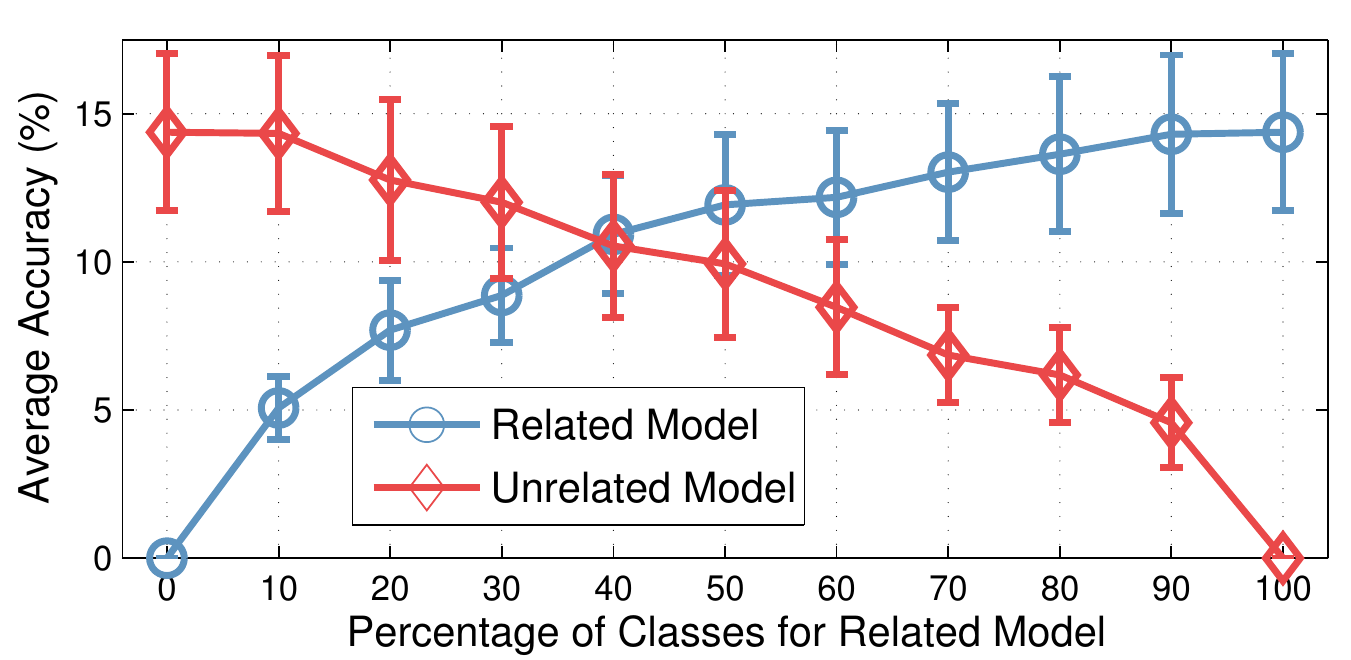}}
\subfigure[Minimal class name affinity]{\includegraphics[width=0.8\linewidth]{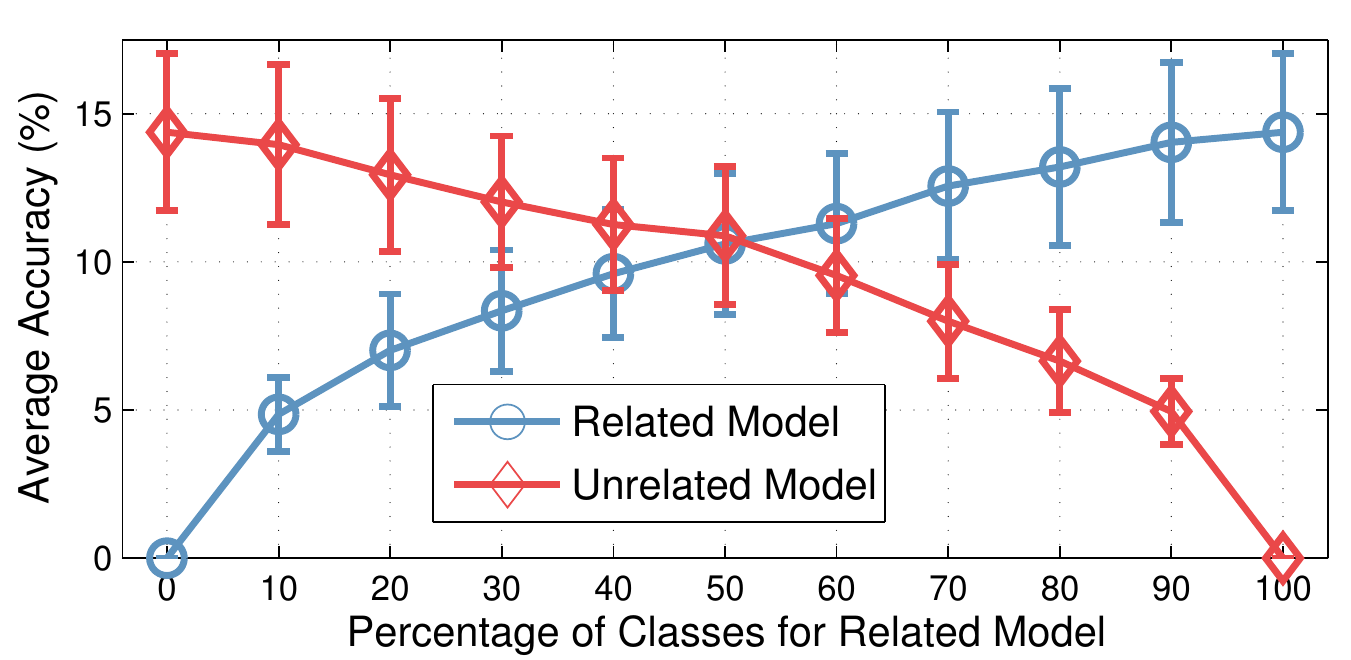}}
\caption{Testing the ability to predict ZSL class transferability by class name affinity: A comparison of models selecting related versus un-related classes as training data.}\label{fig:RelatedModelPerformance}
\end{figure}}

The main observations are as follows: 
(i) A crossover happens at $30\%$ for maximal  class name affinity, which means the model learned on the $30\%$ subset of related training classes outperforms the model learned on the much larger $70\%$ of unrelated classes. 
(ii) The maximal class name affinity is most predicative on the efficacy of zero-shot learning as (1) the crossover point is the left most among all three alternative strategies, and (2) at the equal data point (50\%) the related model most clearly outperforms the unrelated model.
(iii) For maximal affinity, as more classes are included the related
  model increases in performance more rapidly than the unrelated one,
  and saturates after the top $50\%$ are included.  { All these observations together indicate that given limited labelling availability, including training classes that are related to testing ones can benefit ZSL performance (as the crossover is to the left of $50\%$).}


\subsection{Zero-Shot Recognition with Old and New Classes}

\begin{figure}
\centering
\includegraphics[width=0.95\linewidth]{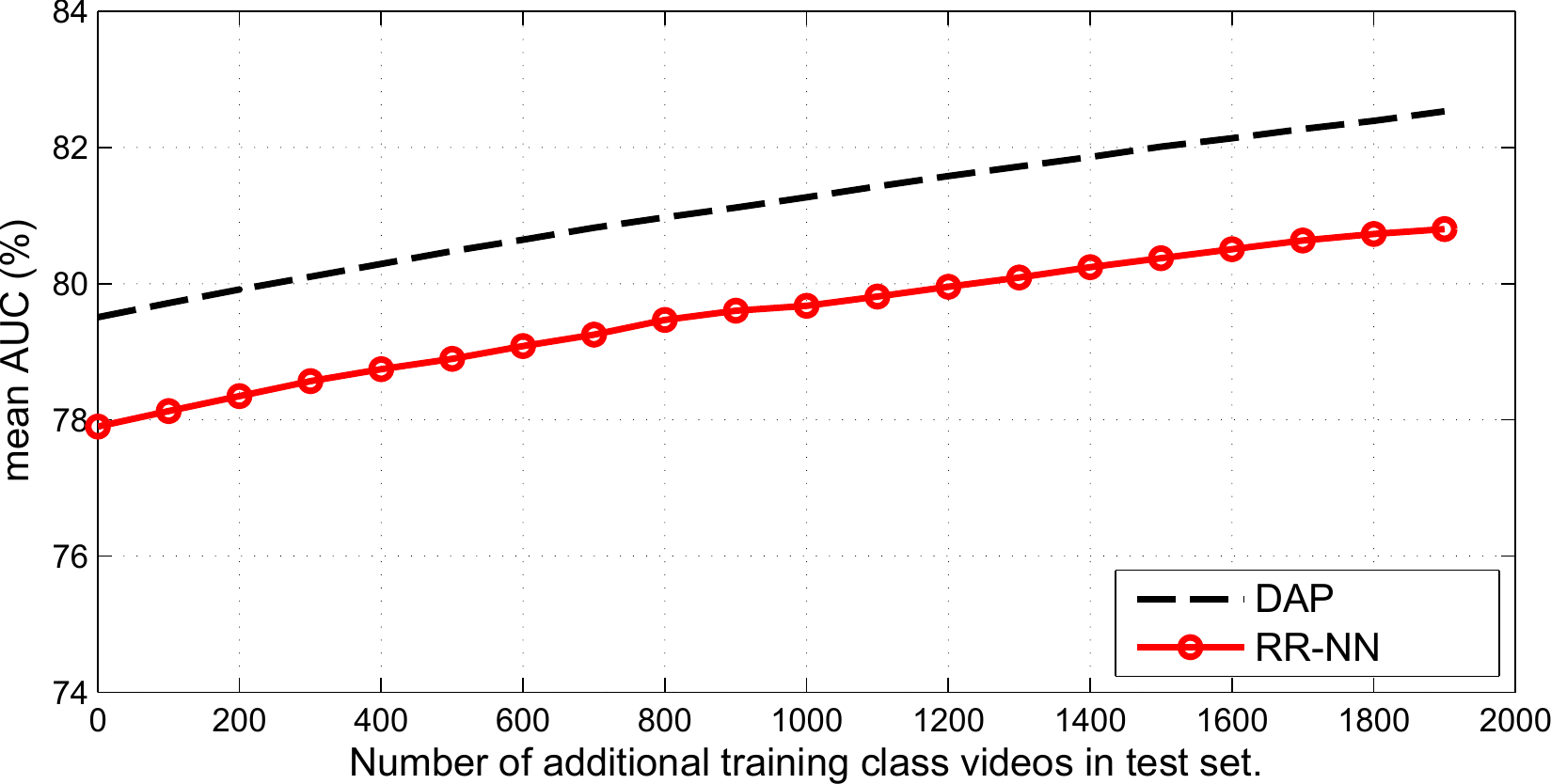}
\caption{Injecting training/known class samples to testing set.}\label{fig:Novelty}
\end{figure}

{Few existing zero-shot learning studies investigate the ability to recognise novel-category visual instances if they occur among known-class instances at testing time. But this may be the setting under which ZSL is used in real applications.  To evaluate how our model performs in the situation where testing instances are mixed with data from training classes, we follow the protocol proposed in \citet{Rohrbach2010}. Specifically, we choose the first data split from UCF101 dataset and hold out 0-1900 training videos evenly from each training/known class for testing. ZSL models are then trained on the reduced training set. In the testing phase, we label all the held-out training videos as negatives of all testing classes and evaluate AUC for each testing class. We compare two models: (1) attribute-based model (DAP) used in \citet{Rohrbach2010}; and (2) our direct similarity based prediction (RR-NN) which corresponds to the final model without data augmentation introduced in Table~\ref{tab:ComponentEval}. By increasing the number of distractor training videos, we observe from Fig.~\ref{fig:Novelty} a steady increase of mean AUC for both attribute-based approach (DAP) and our direct similarity matching (RR-NN). This suggests that both DAP and our model are fairly robust when novel classes must be detected among a background of known classes.}

\subsection{{Imbalanced Test Set}}

\begin{figure}[!h]
\centering
\includegraphics[width=0.99\linewidth]{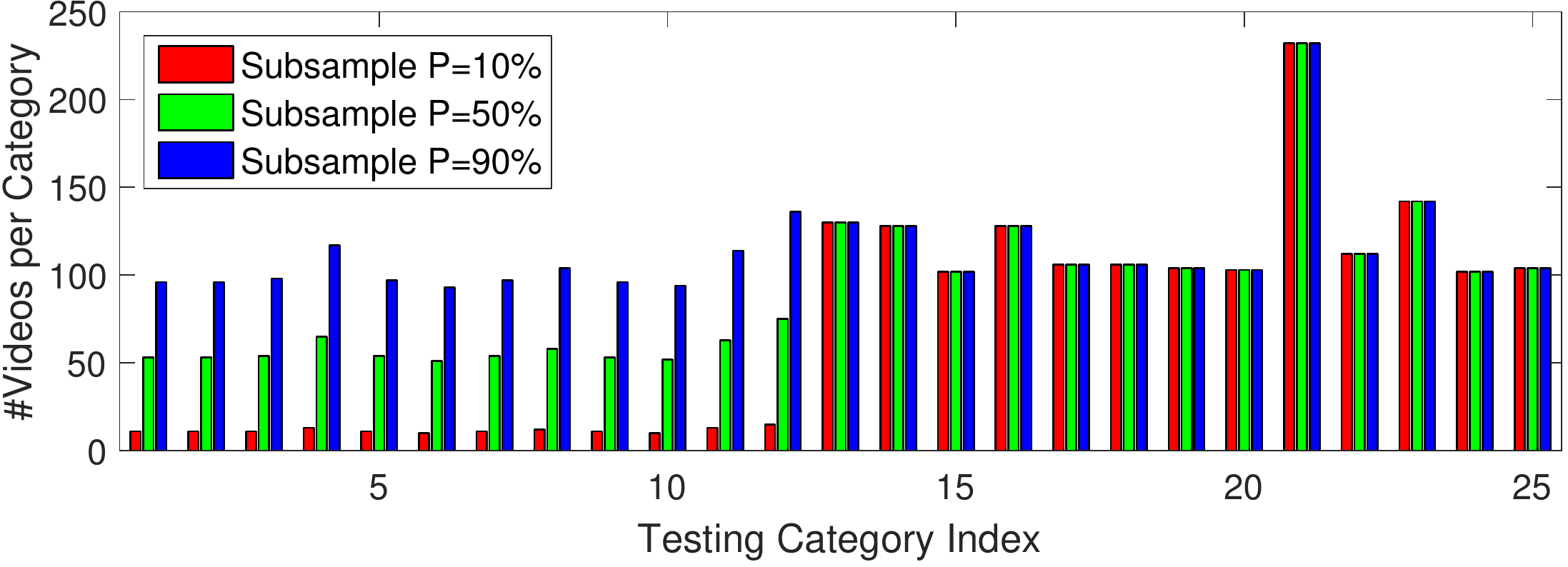}
\caption{Distribution of testing videos after subsampling.}\label{fig:NumVidDist}
\end{figure}

{Transductive strategies have been studied by many existing works \citep{Fu2015,dinu2015improving}, however none of these works have ever studied the assumptions of test set for successful transductive ZSL. In particular, we note that, in zero-shot scenarios, testing categories could be highly imbalanced. How does the transductive strategies generalize to imbalanced test set remains an untouched problem. To verify this aspect, we carry out a particular experiment. Specifically, we experiment on the first split of HMDB51 and randomly subsample $P\%$ testing data from each of the first 12 testing categories for ZSL evaluation. We illustrate the distribution of testing videos per category for $P=10,50,90$ in Fig.~\ref{fig:NumVidDist}.}

\begin{figure}[!h]
\centering
\includegraphics[width=0.95\linewidth]{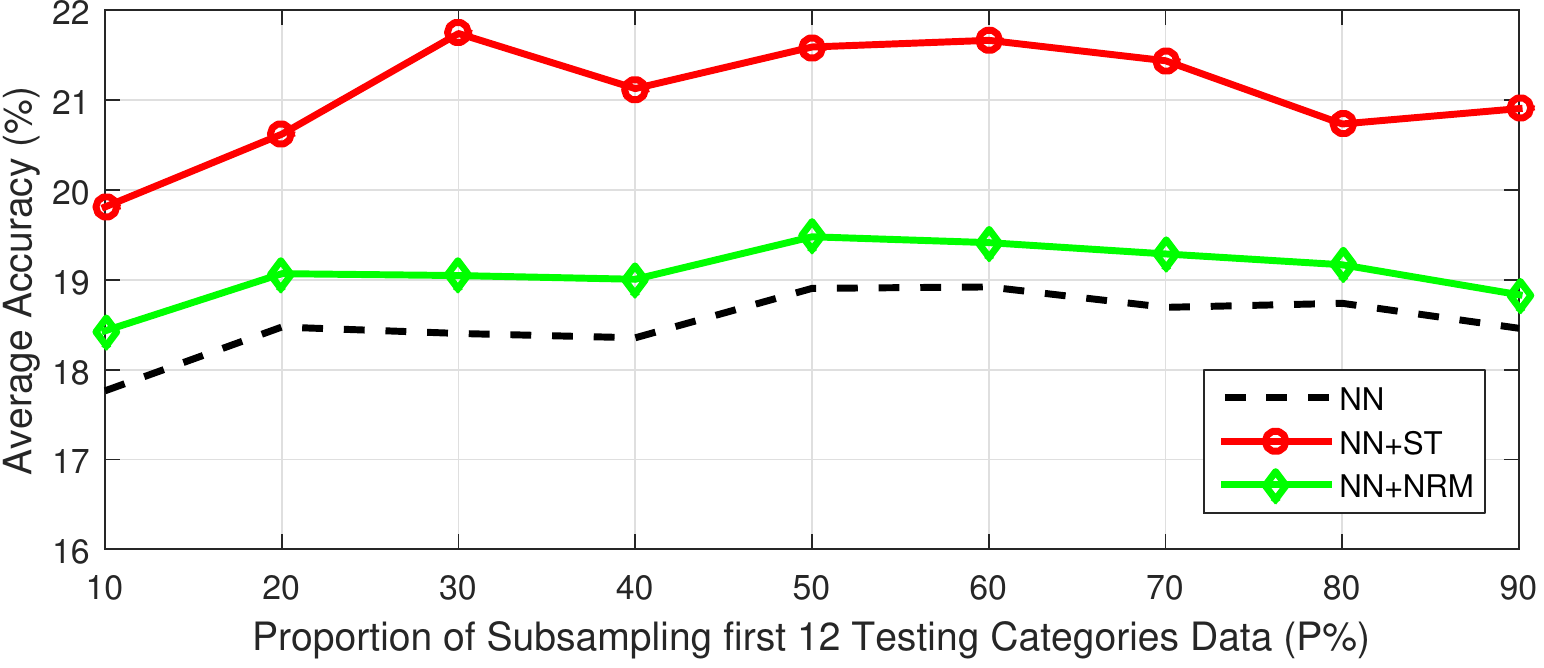}
\caption{Performance of ZSL for subsampled imbalanced test set.}\label{fig:SubSampPerf}
\end{figure}

{Then we experiment the baseline model - NN and two transductive variants - NN+ST and NN+NRM. By increasing $P$ from 10 to 90 we observe from Fig.~\ref{fig:SubSampPerf} that both self-training (red) and hubness correction (green) improve consistently over non-transductive baseline (black dashed). This suggests our transductive strategies are robust to imbalanced test set.}

\subsection{Multi-Shot Learning}\label{sect:MultishotLearning}

We have thus far focused on the efficacy of unsupervised word-vector embeddings for zero-shot learning. In this section we verify that the same representation also performs comparably to state-of-the-art for standard supervised (multi-shot) action recognition. We use the standard data splits and evaluation metrics for all 4 datasets. 

\vspace{0.1cm}\noindent \textbf{Alternatives:}\quad We compare our approach to:
\begin{enumerate}

\item \textbf{Low-Level Feature \citep{Wang2013}} the state-of-the-art results based on low-level features.
\item \textbf{Human-Labelled Attribute (HLA) \citep{Zheng2014}} Exploits an alternative semantic space using human labelled attributes. The model trains binary linear SVM classifiers for attribute detection and uses the vector of attribute scores as a representation. A SVM  classifier with RBF kernel is then trained on attribute representation to predict final labels.
\item \textbf{Data Driven Attribute (DDA) \citep{Zheng2014}} Learns attributes from data using dictionary learning. These attributes are complementary to the human labelled ones. Automatically discovered attributes are processed in the same way as HLA for action recognition.
\item \textbf{Mixed attributes (Mix) \citep{Zheng2014}} A combination of HLA and DDA is applied to exploit the complementary information in two attribute sets.
\item \textbf{Semantic embedding model (Embedding)} first learns a word-vector embedding based on regularized linear regression, as in ZSL. But the standard supervised learning data-split is adopted. All data are mapped into the semantic space via regression and a linear SVM classifier is trained for each category with the mapped training data. 
\end{enumerate}

The resulting accuracies are shown in Table~\ref{tab:MultishotAcc}. We  observe that our semantic embedding  is comparable to the state-of-the-art low-level feature-based classification and is comparable or slightly better than the conventional attribute-based intermediate representations despite the fact that no supervised manual attribute definition and annotation is required.

\begin{table*}[!htb]
\centering
\caption{Standard supervised action recognition. Average accuracy for HMDB51 and UCF101 datasets. Mean average precision for Olympic Sports and CCV.}
\resizebox{0.9\linewidth}{!}{
\begin{tabular}{l|c|c|c|c}
\toprule
\multicolumn{1}{c|}{\bf Method} & {\bf HMDB51} & {\bf UCF101} & {\bf Olympic Sports} & {\bf CCV}  \\ \hline
Low-Level Feature \citep{Wang2013}           & 58.4   & 84.6   & 92.1           & 68.0 \\ \hline
HLA \citep{Zheng2014}                         & -      & 81.7   & -              & -    \\ \hline
DDA \citep{Zheng2014}                        & -      & 79.0   & -              & -    \\ \hline
Mix \citep{Zheng2014}                         & -      & 82.3   & -              & -    \\ \hline
Embedding                   & 56.4   & 82.0   & 93.4           & 51.6 \\ \bottomrule
\end{tabular}  \label{tab:MultishotAcc}%
}
\end{table*}

\subsection{{Efficiency and Runtime}}

{Our ZSL algorithm  is easy to implement and has favourable efficiency. We estimate the computation complexity for solving manifold regularized regression in Eq~(\ref{eq:ManifoldCloseform}) to be $O(2N^3+d_zN)$ (assume the schoolbook matrix multiplication algorithm). Nevertheless, if the number of training data $N$ is too large to fit into memory, our model can be solved by stochastic gradient descent (SGD). The gradient w.r.t. mapping $\matr{A}$ is}

{
\begin{equation}
\begin{split}
\nabla\matr{A}=&\frac{1}{n_l}\left(-2\tilde{\matr{z}_i}\matr{k}_i^\top\matr{J}+2\matr{A}\matr{k}_i\matr{J}\matr{k}_i^\top\right)\\
&+2\gamma_A\matr{A}\matr{k}_i+\frac{\gamma_I}{(n_l+n_u)^2}2\matr{A}\matr{k}_i\matr{l}_i^\top\matr{K}^\top
\end{split}
\end{equation}
}

\noindent {
for which we estimate the computation complexity for each iteration to be $O(4d_z+N^2)$. }

{In our implementation, it
  takes about 300 seconds (including overhead) to train and test on 50 splits of the entire HMDB51
  benchmark dataset (6766 videos of 51 categories of actions), or 520
  seconds with data augmentation, using a server with 32 Intel E5-2680 cores. The runtime is dominated
by the matrix inversion in Eq.~(\ref{eq:ManifoldCloseform}).}

\section{Detailed Parameter Sensitivity Analysis}
\textcolor{black}{In the main experiments we set the free parameters ridge regularizor $\gamma_A=10^{-6}$, manifold regularizor $\gamma_I=40$, manifold Knn graph $N^{G}_K=5$, Self-Training Knn $N^{st}_K=100$.} In this section  we analyse the impact of these free parameters in our model. 

\subsection{word-vector Dimension}
\begin{figure}[!htb]
\centering
\includegraphics[width=.95\linewidth]{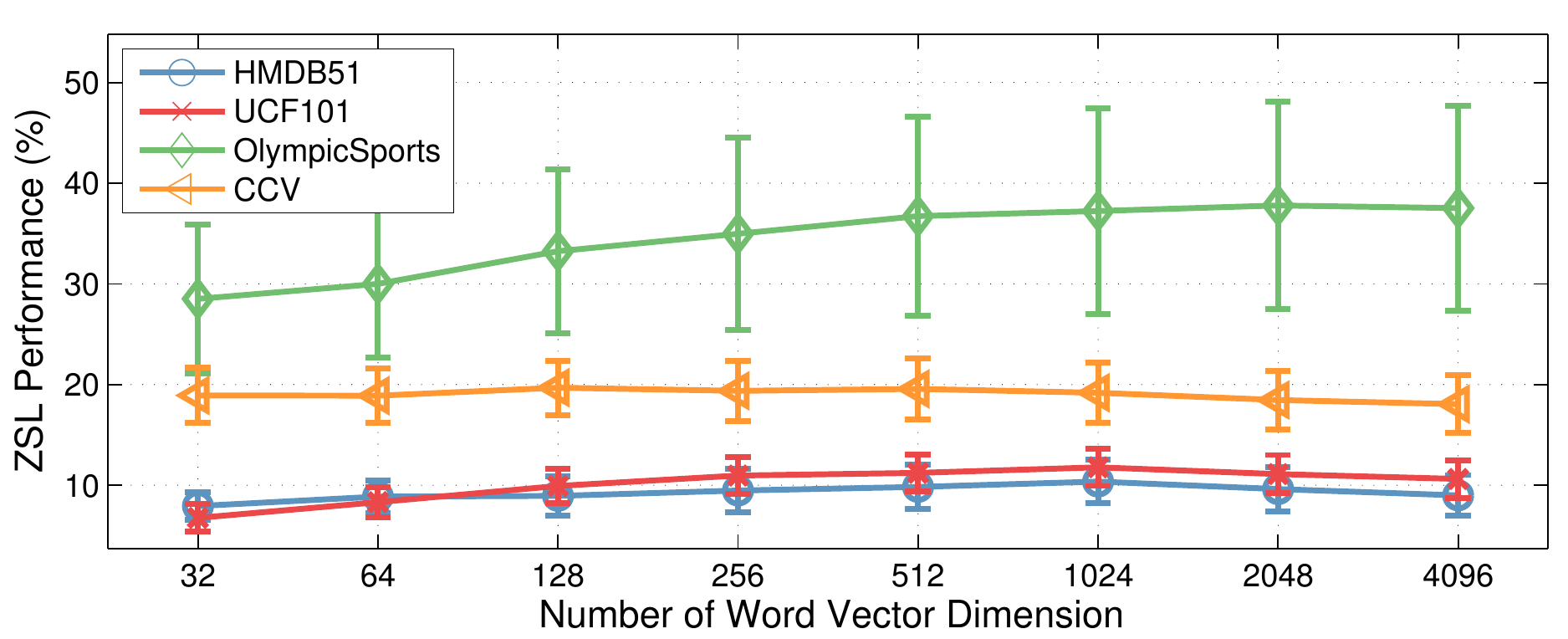}
\caption{Zero-shot performance v.s. dimension of word-vector.}\label{fig:embedding_dim}
\end{figure}

We investigate how  the specific word-vector model $\vect{z}=g(\scal{y})$ affects the performance of our framework.
For the study of word-vector dimension we train word-vectors on 4.6M Wikipedia documents\footnote{Google News Dataset is not publicly accessible. So we use a smaller but public dataset - 4.6M Wikipedia documents.} and vary dimension from 32 to 1024. We then evaluate the performance of zero-shot and multishot learning v.s. different dimension of embedding space. The results are given in Fig.~\ref{fig:embedding_dim}. 

We  observe that word-vector dimension does affect the zero-shot recognition performance. Performance generally increases with dimension of word-vector from 32 to 4096 in HMDB51, UCF101 and Olympic Sports, while showing no clear trend for CCV. In general a reasonable word-vector dimension is between 256 to 2048. 

\subsection{Visual to Semantic Mapping}
\begin{figure}[!h]
\centering
\includegraphics[width=0.95\linewidth]{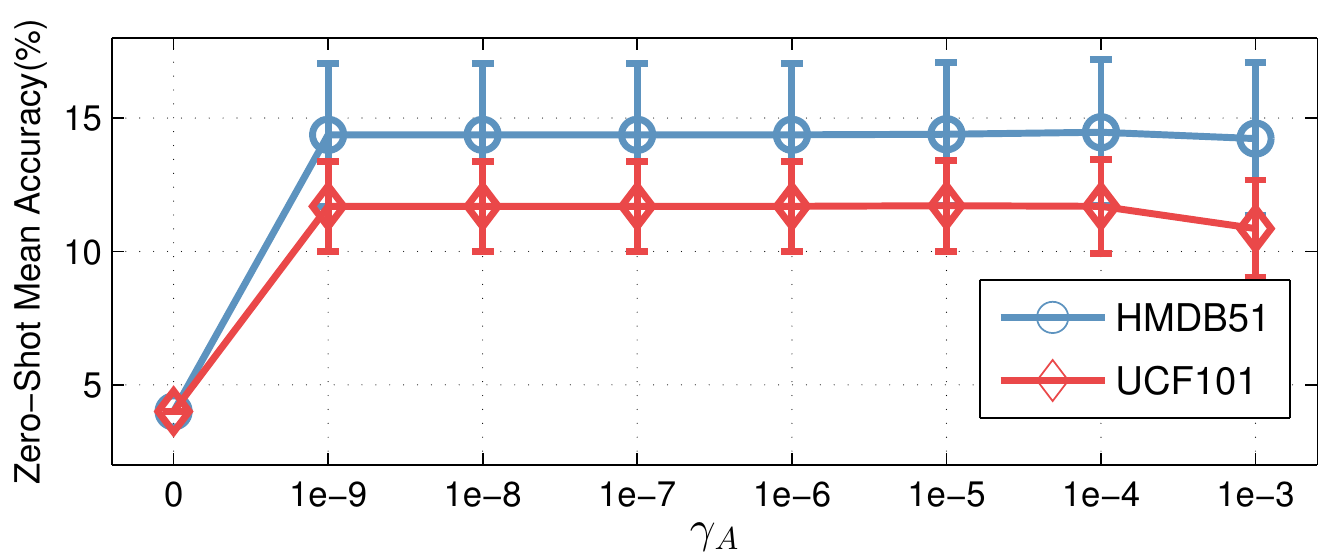}
\caption{Zero-shot mean accuracy v.s. ridge regression parameter}\label{fig:RidgeParameter}
\end{figure}

\noindent\textbf{Ridge regression regularization:}\quad We learn the visual to semantic mapping with regularized linear regression. The regularization parameter $\gamma_A$ controls the regression model complexity. Here, we study the impact of $\gamma_A$ on zero-shot performance. We measure the 50 splits' average accuracy by varying $\gamma_A$ in the range of 
$\{0,10^{-9},10^{-8},\cdots,10^{-3}\}$. A plot of zero-shot mean accuracy v.s. regularization parameter is given in Fig.~\ref{fig:RidgeParameter}. From this figure we observe that our model is insensitive to the ridge parameter for any non-zero regularizer. {However, when no regularization is used the performance is close to random. This is due to all zero or co-linear rows/columns in the kernel matrix which causes numerical problems in computing the inverse.}

\noindent\textbf{Manifold regression:}\quad
{We have seen that transductively exploiting
  testing/unlabelled data in manifold learning improves zero-shot
  performance}. Two parameters are involved: the manifold
regularization parameter $\gamma_I$ in Loss function
(Eq.~\ref{eq:ManifoldRegressionLossFcn}) and $N^{G}_K$ in constructing
the symmetrical KNN graph. $\gamma_I$ controls the preference for
preserving the manifold structure in mapping to the semantic space,
versus exactly fitting the training data. Parameter $N^{G}_K$
determines the precision in modelling the manifold structure. Small
$N^{G}_K$ may more precisely exploit the testing data manifold,
however it is more prone to noise in the neighbours. 

\begin{figure}
\centering
\subfigure[HMDB51]{\includegraphics[width=0.78\linewidth]{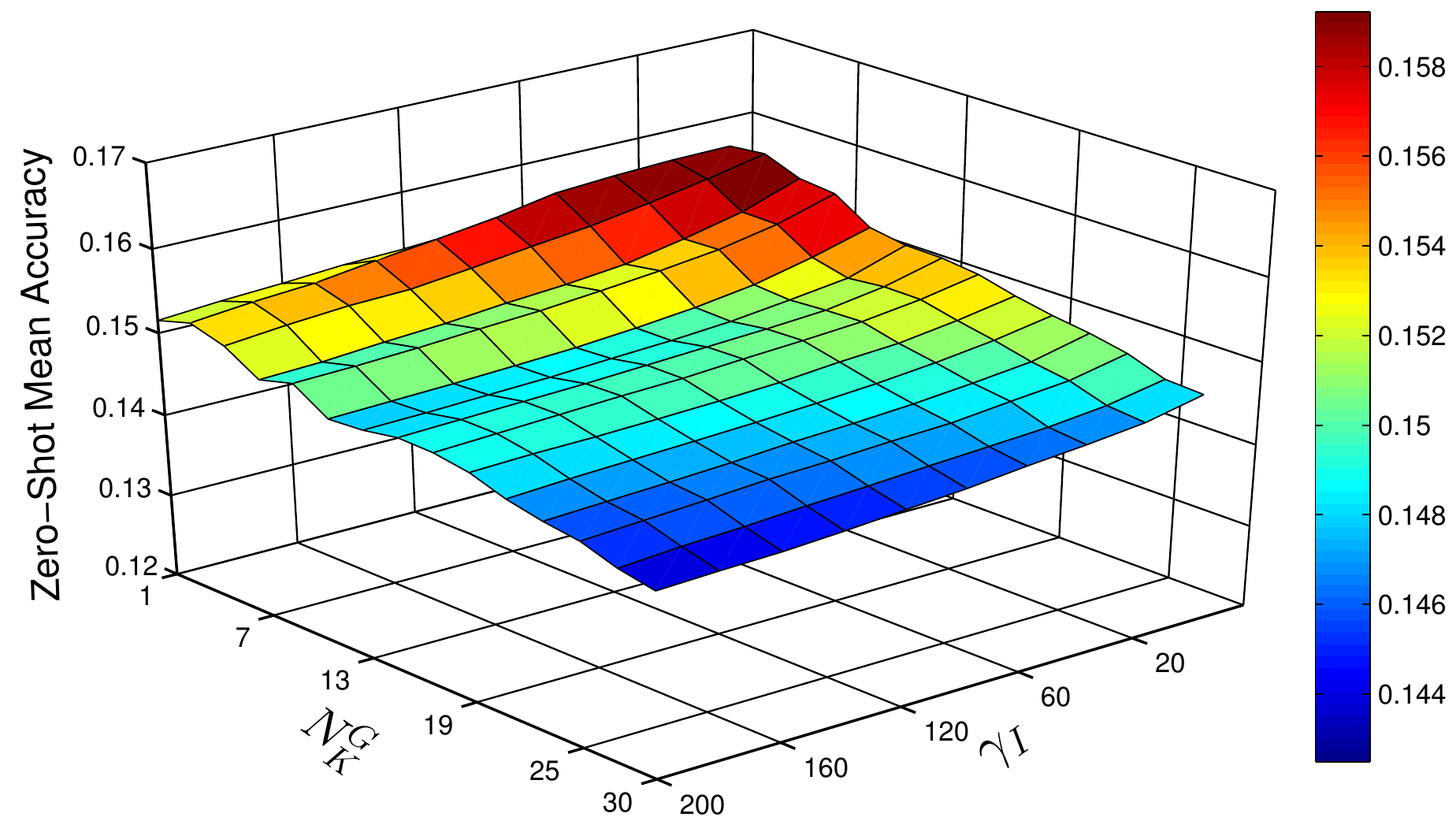}}\vspace{-0.3cm}\\
\subfigure[UCF101]{\includegraphics[width=0.78\linewidth]{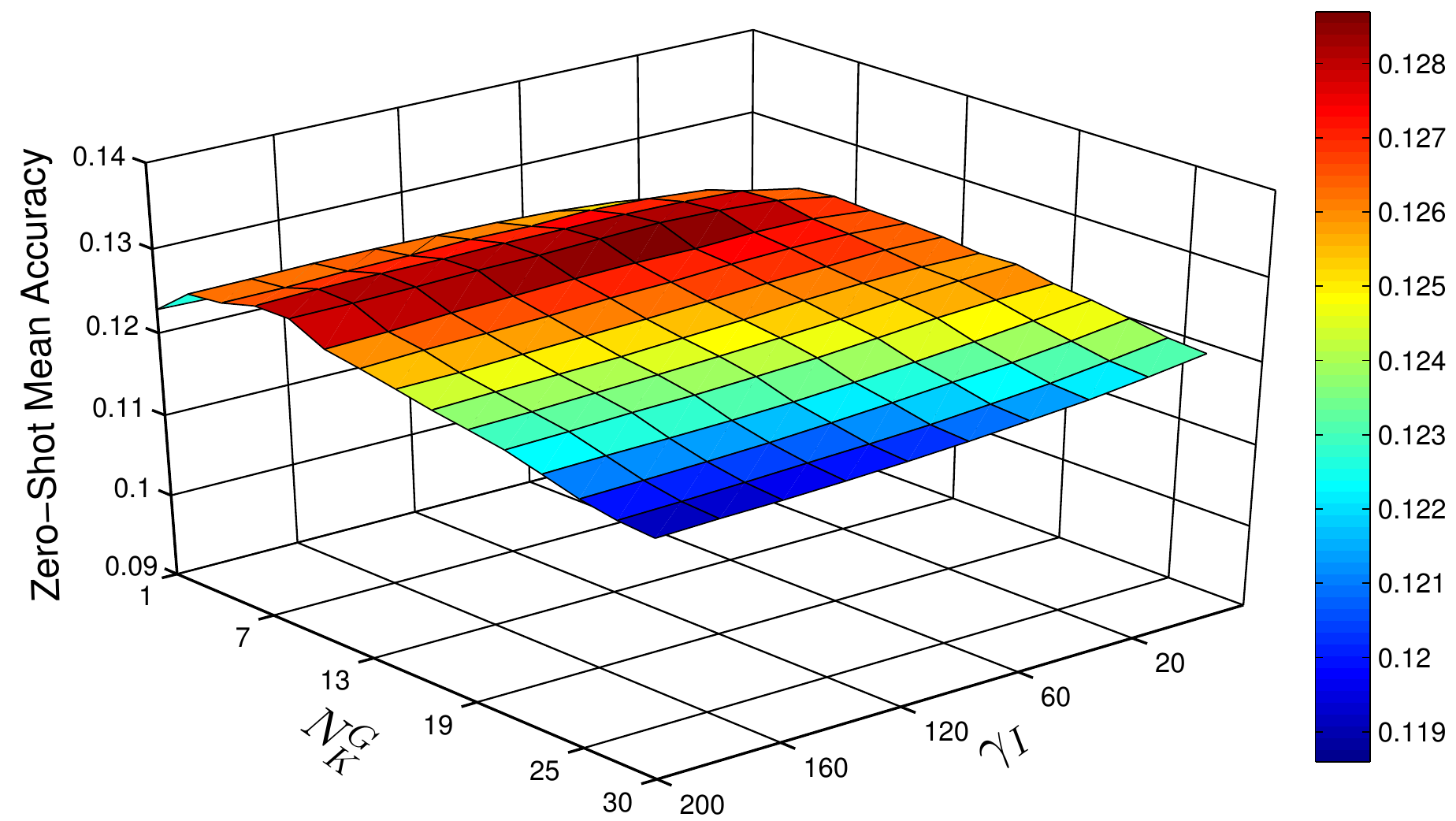}}\vspace{-0.3cm}
\caption{Zero-shot recognition accuracy with respect to manifold regression parameters $\gamma_I$ and $N^{G}_K$.}\vspace{-0.2cm}\label{fig:paramKI}
\end{figure}

Here we analyse the impact of these two parameters, $\gamma_I$ and $N^{G}_K$ by measuring  zero-shot recognition accuracy on HMDB51 and UCF101. We evaluate the joint effect of $\gamma_I$ and $N^{G}_K$ while fixing $\gamma_A=10^{-6}$. Specifically we test $\gamma_I\in\{20,40,\cdots,100\}$ and $N^{G}_K\in\{1, 3, 5 \cdots, 29\}$. The results in Fig.~\ref{fig:paramKI} show that there is a slightly preference towards moderately low values of $N^{G}_K$ and $\gamma_I$, but the framework is not very sensitive to these parameters.

\subsection{Self-Training}
We previously demonstrated in Table~\ref{tab:ComponentEval}, that
self-training  (Section~\ref{sect:MulZeroLearning}) helps to mitigate
the domain shift problem. Here, we study the influence of the
$N^{st}_K$ parameter for KNN in self-training. Note the $N^{st}_K$
concerns the neighbouring data distribution around prototypes at
testing time rather than manifold regularization KNN graph $N^{G}_K$
at training time. We evaluate \newline $N^{st}_K \in \{ 1, 2, 3,
\cdots, 200 \}$. To thoroughly examine the effectiveness of
self-training, we investigate all baselines with self-training
introduced in section \ref{sect:exp_ZSL} including 

\noindent \begin{itemize}
\item {\text{\sffamily X}-RR-\text{\checkmark}-NN-\text{\checkmark}}\quad\quad\quad(\textbf{NN+ST})
\item {\text{\sffamily X}-RR-\text{\checkmark}-NRM-\text{\checkmark}}\quad\quad({\bf NRM+ST})
\item {\text{\sffamily X}-RR-\text{\checkmark}-GC-\text{\checkmark}}\quad\quad\quad({\bf GC+ST})
\item {\text{\sffamily X}-MR-\text{\checkmark}-NN-\text{\checkmark}}\quad\quad\quad({\bf Manifold+ST})
\item {\text{\sffamily X}-MR-\text{\checkmark}-NRM-\text{\checkmark}}\quad\quad({\bf Manifold+NRM+ST})
\item {\text{\sffamily X}-MR-\text{\checkmark}-NRM-\text{\checkmark}}\quad\quad({\bf Manifold+NRM+ST})
\item {\text{\checkmark}-RR-\text{\checkmark}-NN-\text{\checkmark}}\quad\quad\quad({\bf NN+Aux+ST})
\item {\text{\checkmark}-RR-\text{\checkmark}-NRM-\text{\checkmark}}\quad\quad({\bf NRM+Aux+ST})
\end{itemize}

\begin{figure}[!h]
\centering
\subfigure[HMDB51]{\includegraphics[width=0.49\linewidth]{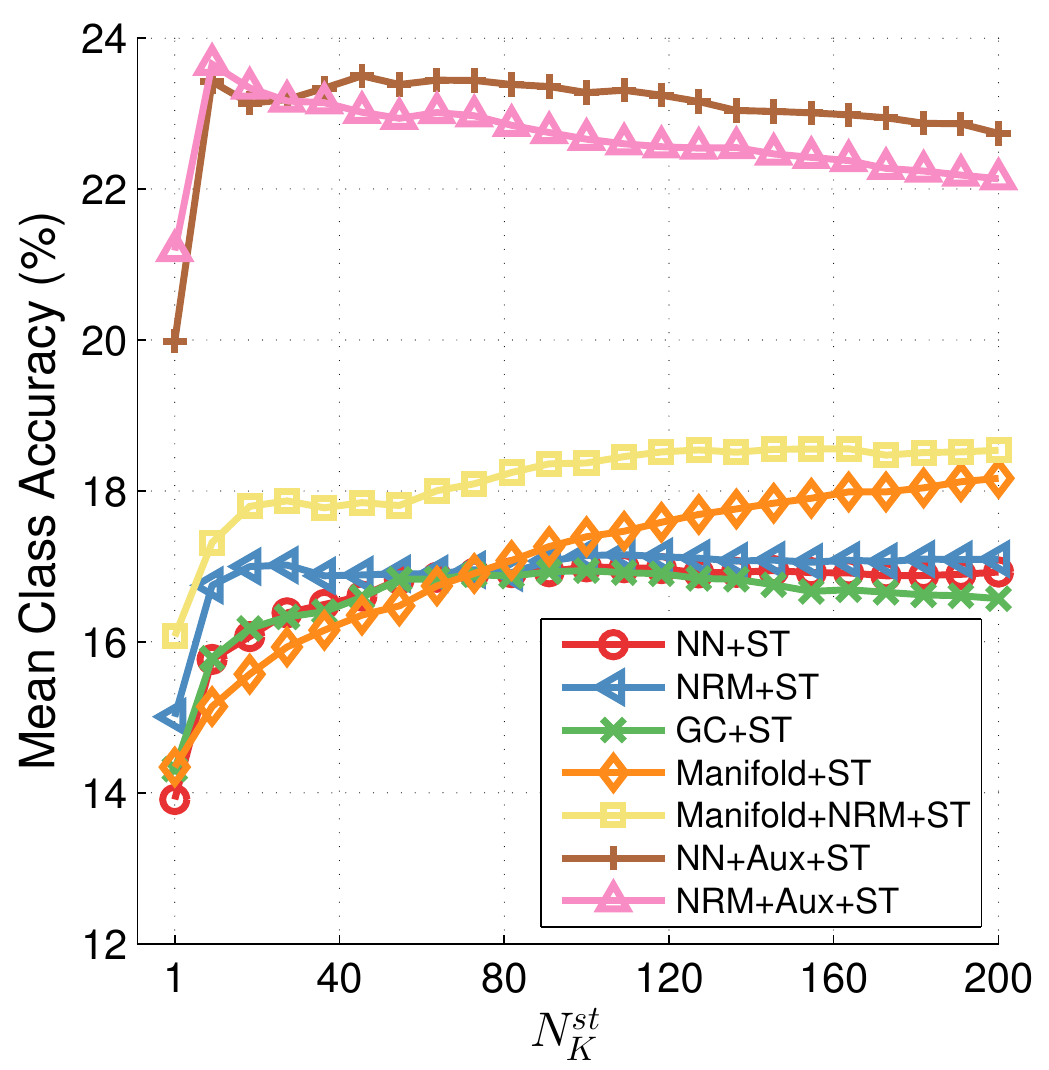}}
\subfigure[UCF101]{\includegraphics[width=0.49\linewidth]{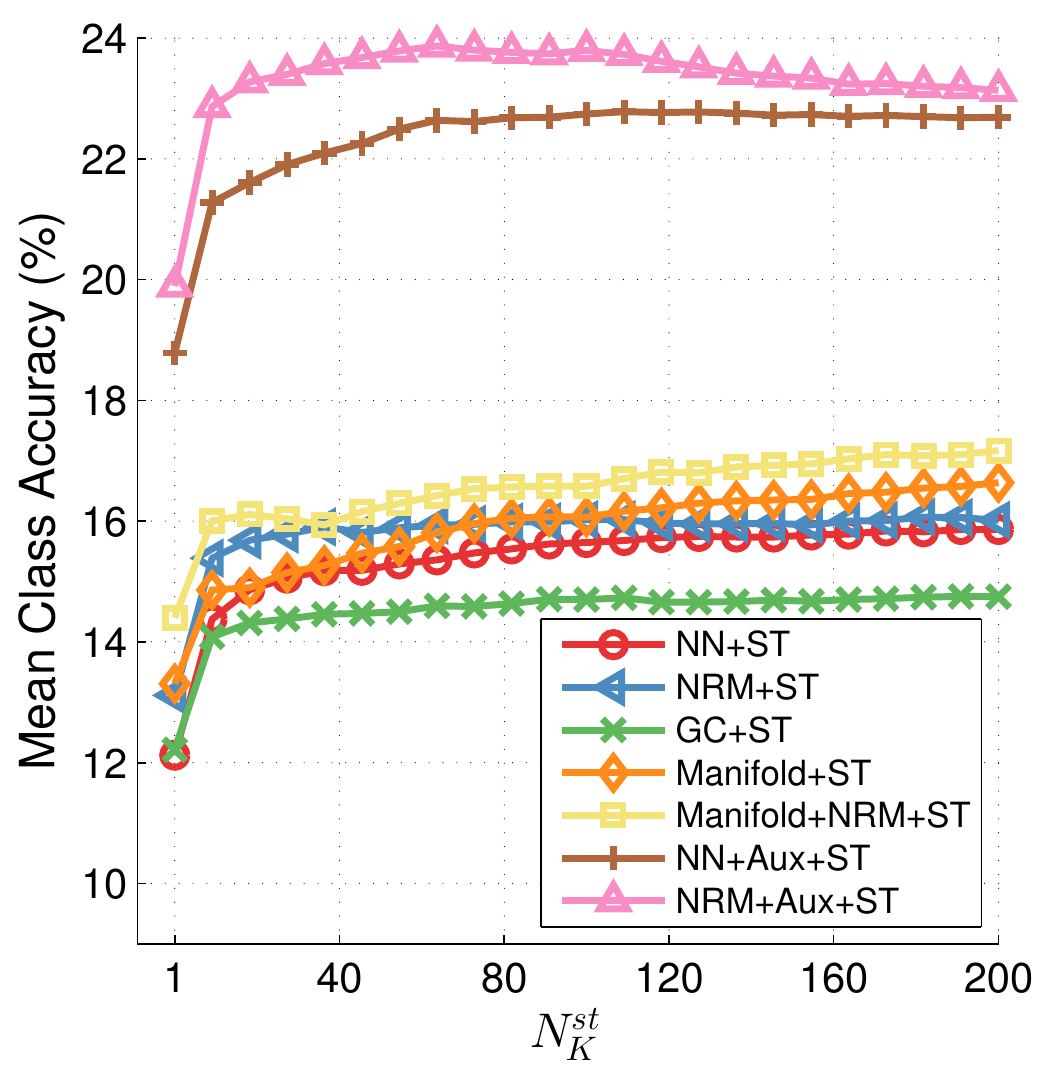}}
\caption{Zero-shot recognition accuracy v.s. self-training parameter K.}\label{fig:ST_K_Eval}
\vspace{-0.2cm}
\end{figure}

The accuracy v.s. $N^{st}_K$ is illustrated in Fig.~\ref{fig:ST_K_Eval}. Performance is robust to $N^{st}_K$ when $N^{st}_K$ is above 20.

\section{Conclusion}
In this study, we investigated {\em unsupervised} word-vector embedding
space representation for zero-shot action recognition for the first
time. The fundamental challenge of zero-shot learning is the disjoint
training and testing classes, and associated domain-shift. We explored
the impact of four simple but effective strategies to address this: data
augmentation, manifold regularization, self-training and hubness
correction. Overall we demonstrated that given auxiliary and transductive access to testing data these strategies are
complementary, and together facilitate a highly effective system that
{ is even competitive against existing attribute-based approaches. If manually labelled attributes are available,  our transductive strategies can produce the state-of-the-art performance.} {Moreover, our
model has a closed-form solution that is very simple to implement (a few lines of matlab) and runs very efficiently}. Finally, we also provide a unique analysis of the inter-class
affinity for ZSL, giving insight into why and when ZSL
works. This provides for the first time two new capabilities: the
ability to predict the efficacy of a given ZSL scenario in advance,
and a mechanism to guide the construction of suitable training sets for a
desired set of target classes. 

{
We have done some preliminary investigation of recognising novel classes when testing instances also include those of known training classes -- a setting which is practically valuable but little studied. Designing algorithms specifically to deal with this challenging setting has received limited attention  \citep{Socher2013}, and is still an open question.
}
{
Another issue which is not fully addressed in this work is transferability prediction. Given limited labelling ability, it is desirable to annotate most useful training data to support zero-shot recognition. We discussed one possible way --  measuring the semantic relatedness between candidate training class and testing classes. However the relation could be more complicated than pairwise, and the inclusion of a new training class could affect recognition of unknown classes in together with other training classes. How to best utilise labelling effort to support zero-shot recognition remains an open question.
}


\bibliographystyle{spbasic}      
\bibliography{./IJCV}   

%
%

\end{document}